\documentclass[11pt]{article}
\usepackage[utf8]{inputenc}
\usepackage{authblk}
\usepackage{amsthm}
\usepackage{amsmath}
\usepackage{amssymb}
\usepackage{comment}
\usepackage{graphicx}
\usepackage{graphics}
\usepackage{booktabs}
\usepackage{subfig}
\usepackage{color}
\usepackage{hyperref}
\usepackage[left = 1in, right = 1in]{geometry}

\usepackage{cite}
\newtheorem{definition}{Definition}

\title{A Causal Lens for Peeking into Black Box Predictive Models: Predictive Model Interpretation via Causal Attribution}
\author{
Aria Khademi,\textsuperscript{\rm1,\rm2}
Vasant Honavar\textsuperscript{\rm1,\rm2,\rm3,\rm4}\\
\textsuperscript{\rm1} Artificial Intelligence Research Laboratory\\
\textsuperscript{\rm2} College of Information Sciences and Technology\\
\textsuperscript{\rm3} Department of Computer Science and Engineering\\
\textsuperscript{\rm4} Institute of Computational and Data Sciences\\
The Pennsylvania State University\\
\{khademi,vhonavar\}@psu.edu
}

\begin{document}

\maketitle

\begin{abstract}
    With the increasing adoption of predictive models trained using machine learning across a wide range of high-stakes applications, e.g., health care, security, criminal justice, finance, and education, there is a growing need for effective techniques for explaining such models and their predictions. We aim to address this problem in settings where the predictive model is a {\em black box}; That is, we can only observe the response of the model to various inputs, but have no knowledge about the internal structure of the predictive model, its parameters, the objective function, and the algorithm used to optimize the model. We reduce the problem of interpreting a black box predictive model to that of estimating the {\em causal effects} of each of the model inputs on the model output, from observations of the model inputs and the corresponding outputs. We estimate the causal effects of model inputs on model output using variants of the Rubin Neyman {\em potential outcomes} framework for estimating causal effects from observational data. We show how the resulting {\em causal attribution} of responsibility for model output to the different model inputs can be used to interpret  the predictive model and to explain its predictions. We present results of experiments that demonstrate the effectiveness of our approach to the interpretation of black box predictive models via causal attribution in the case of deep neural network models trained on one synthetic data set (where the input variables that impact the  output variable are known by design) and two real-world data sets: Handwritten digit  classification, and Parkinson's disease severity prediction. Because our approach does not require knowledge about the predictive model algorithm and is free of assumptions regarding the black box predictive model except that its input-output responses be observable, it can be applied, in principle, to {\em any} black box predictive model.

\end{abstract}


\section{Introduction}\label{intro}
Our ability to acquire and annotate increasingly large amounts of data together with rapid advances in machine learning have made predictive models that are trained using machine learning ubiquitous in virtually all areas of human endeavor. Recent years have seen rapid adoption of machine learning to automate decision making in many high-stakes applications such as health care \cite{norgeot2019call,wang2019deep,wiens2018machine,rajkomar2019machine,esteva2019guide,leung2015machine,miotto2018deep}, finance \cite{sirignano2016deep,albanesi2019predicting,jagtiani2019roles,croux2020important}, criminal justice \cite{pelzer2018policing}, security \cite{horowitz2018artificial,tripathi2018suspicious}, education \cite{waters2014grade,larrabee2019efficacy,romero2010educational}, and scientific discovery \cite{mjolsness2001machine,butler2018machine,dunjko2018machine,camacho2018next,vu2018shared,gil2014amplify}. In  such high-stakes applications, the predictive models, produced by the state-of-the-art machine learning algorithms, e.g., deep learning \cite{lecun2015deep}, kernel methods \cite{hofmann2008kernel}, among others,  are often complex black boxes that are hard to explain to users \cite{burrell2016machine,gunning2017explainable}.

\noindent{\bf Example.} Consider a medical decision making scenario where a predictive model, e.g., a deep neural network, trained on a large database of labeled data, is to assist physicians in diagnosing patients. Given the high-stakes nature of the application, it is important that the clinical decision support system be able to explain the output of the deep neural network to the physician, who may not have a deep understanding of machine learning. For example, the physician might want to understand the subset of patient characteristics that contribute to the diagnosis; or the reason as to why  diagnoses were different for two different patients, etc.

In high-stakes applications of machine learning,  the ability to explain a machine learned predictive model is a prerequisite for establishing {\em trust} in the model's predictions. Hence, there is a growing interest in machine learning algorithms that produce {\em interpretable} (as opposed to black box) models as well as techniques for interpreting black box models \cite{lipton2016mythos,doshi2017towards,gunning2017explainable}. 

\subsection{Model Explanation and Interpretation}
The nature and desiderata of explanations have been topics of extensive study in philosophy of science, cognitive science, and social sciences \cite{kitcher1987van,salmon1984scientific,salmon1998causality,kass1988need,miller2019explanation}. Arguably, predictive model {\em interpretation} is a necessary condition for model {\em explanation}. A key tool for model interpretation is {\em attribution} of responsibility for the model output to the model's inputs (i.e., features) \cite{sundararajan2017axiomatic, roscher2020explainable}. Hence, a large body of work has focused on methods for  interpretation of black box predictive models (reviewed in \cite{mueller2019explanation,guidotti2018survey,adadi2018peeking,montavon2018methods}), also known as {\em posthoc} interpretations \cite{guidotti2018survey,lipton2016mythos}. They include methods for visualizing the effect of the model inputs on its outputs \cite{simonyan2013deep,yosinski2015understanding,zeiler2014visualizing}, methods for extracting purportedly human interpretable rules from black box models \cite{andrews1995survey,frosst2017distilling,setiono1995understanding,towell1993extracting,thrun1995extracting,letham2015interpretable}, feature scoring methods  that assess the importance of individual features on the prediction \cite{friedman2001greedy,goldstein2015peeking,datta2016algorithmic,lipovetsky2001analysis,vstrumbelj2014explaining,lundberg2017unified,ancona2019explaining,chen2018learning}, gradient based methods  that assess how changes in inputs impact the model predictions \cite{baehrens2010explain,shrikumar2016not,shrikumar2017learning,bach2015pixel,chen2018learning}, and techniques for approximating local decision surfaces in the neighborhood of the input sample via localized regression \cite{ribeiro2016should,selvaraju2017grad,bhatt2020explainable}. A shared feature of all of these model interpretation methods is that they primarily focus on how a model's inputs correlate with its outputs. As has been pointed out, they often fail to generate reliable attributions \cite{sundararajan2017axiomatic,kindermans2019reliability}, let alone interpretations that support explanations \cite{mueller2019explanation,guidotti2018survey,adadi2018peeking,montavon2018methods}. 

\subsection{Causal Underpinnings of Model Explanation and Interpretation}
Satisfactory explanations  have to provide answers to questions such as  ``What features of the input are responsible for the  predictions?'';  ``Why are the model's outputs different for two individuals?'' (e.g., ``Why did John's loan application get approved when Sarah's was not?''). As Salmon noted, ``We come to understand a phenomenon when we can explain {\em why}  it occurred \cite{salmon1998causality} (pp. 83)'', where  ``why'', has a {\em causal} meaning. In the words of Halpern and Pearl  \cite{halpern2005causes}, ``the role of explanation is to provide the information needed to establish {\em causation} '' (emphasis ours) \cite{halpern2005causes}. Hence, satisfactory explanations are fundamentally {\em causal} in nature \cite{salmon1984scientific,salmon1998causality,kitcher1987van,kass1988need,mittelstadt2019explaining,mueller2019explanation,miller2019explanation}. Therefore, interpretations or attributions that assign responsibility for the model's output to the model's inputs, have to necessarily be causal in nature. Purely correlation based methods fail to provide causal attributions, especially in the presence of {\em confounders} (that is, inputs to the predictive model that causally influence both some of the other model inputs as well as model outputs). Establishing causal effect of one variable on another has to effectively cope with confounders \cite{pearl2009causality}. Causal attribution is no exception.  

\subsection{Predictive Model Interpretation via Causal Attribution} 
Recently, \cite{chattopadhyay2019neural} offered the first causal approach for interpretation of the output of a deep neural network in terms of its inputs. They offer a method for {\em causal attribution}, namely, assigning responsibility for the deep neural network output among its inputs.  This is accomplished by estimating the causal effect of each of the model inputs on the model output. They first translate a learned deep neural network model into a {\em functionally equivalent} Structural Causal Model (SCM) or a Causal Bayesian Network  \cite{pearl2009causality} and then use the resulting Causal Bayesian Network to estimate the relevant causal effects, using variants of standard methods for causal inference using Causal Bayesian Networks.

\subsection {Overview and Key Contributions}

The only existing method \cite{chattopadhyay2019neural} for causal attribution of black box predictive models suffers from at least two key limitations:
(i) Since the method relies on a specific translation of a deep neural network into a Causal Bayesian Network, the method is limited in its applicability to black box predictive models that are not deep neural networks; (ii) The method requires the attribution algorithm to have access to the {\em structure} as well as parameters of the deep neural network. In many application scenarios, the users of the predictive model or the causal attribution algorithm lack knowledge of the internal structure of the model and its parameters, the objective function, and the algorithm used to optimize the predictive model. In such cases, the users can only observe the outputs of the model for user-supplied inputs. This further limits the applicability of the causal attribution method in \cite{chattopadhyay2019neural} to settings where the deep neural network model, although complex, is {\em not} a black box.  Against this background, we consider the causal attribution of black box predictive models. 

We note that for a given deep neural network, or for that matter, any predictive model that is trained on a specific training set, a Causal Bayesian Network (CBN) that is functionally equivalent to the trained predictive model simply {\em cannot} include any information that is not already encoded by the trained predictive model.  Hence, we conjecture that it should be possible to recover all of the information encoded by such a Causal Bayesian Network, by observing the outputs of the corresponding predictive model on a sufficiently large sample of inputs. Thus, any causal attributions that can be estimated from the Causal Bayesian Network that is functionally equivalent to a given predictive model, can be equally well estimated from observing the outputs of the predictive model on a sufficiently large sample of inputs. If this conjecture turns out to be true, then it must be possible to leverage the state-of-the-art methods for estimating causal effects from observational data to produce causal explanations of black box predictive models and their predictions.

We reduce the model interpretation  question, ``Why did the predictive model generate the output $Y$ for input $X$?'', to  the following  equivalent question: ``How are the features of the model input $X$ causally related to  the model output $Y$?'' In other words, we reduce the task of interpreting a black box predictive model  to the task of estimating, from observations of the inputs and the corresponding outputs of the model, the causal effect of each input variable or feature on the output variable. We estimate the relevant causal effects under rather mild (and quite standard) assumptions of the  {\em Potential Outcomes} framework \cite{rubin2005causal,hernan2020causal} for estimating causal effects from observational data. Because unlike the only existing causal attribution method \cite{chattopadhyay2019neural}, we do not require the causal attribution method to have access to  structure or parameters of the black box predictive model, the resulting causal attribution method can be applied, in principle, to {\em any} black box predictive model, so long as it can probe the model and  observe the model's response to any supplied input data sample. 

We demonstrate the effectiveness of the proposed approach to interpretation (via causal attribution) of black box predictive models, specifically, deep neural networks (DNN), using state-of-the-art methods for estimating causal effects from observational data, where the input variables are continuous.

The key contributions of this paper are as follows:
\begin{enumerate} 
    \item  We offer the first model agnostic approach to interpretation of black box predictive models via causal attribution, that is, estimation of the causal effect of each of the model's inputs on the model's output. 
    \item We reduce the problem of interpretation of black box predictive models to the well-known problem of estimating the causal effects (of the model's inputs on the model's outputs) from observational data. 
    \item In contrast to the only existing approach to interpretation via causal attribution \cite{chattopadhyay2019neural}, our solution does not require the interpretation algorithm to have access to the internal structure and parameters of the black box predictive model, and hence can be applied, in principle, to {\em any} black box predictive model, so long as it can probe the model and  observe the model's response to any supplied input data sample. 
    \item We show how to use the resulting causal attributions to explain the observed differences in the model's outputs in different cases, e.g., ``Why did the model recommend that John's loan application be approved when Sarah's was not?'' 
    \item We demonstrate the effectiveness of our approach to the interpretation of black box predictive models  via causal attribution  using DNN models trained on one synthetic data set (where the input variables that impact the  output variable are known by design) and two real-world data sets: Handwritten digit  classification, and Parkinson's disease severity estimation. 
\end{enumerate}

The rest of the paper is organized as follows. 
Section \ref{prelim} introduces  the key definitions and the basic machinery of causal inference from observational data which we will utilize in the rest of the paper.  Section \ref{causal-attribution} introduces our approach to interpretation of black box predictive models via causal attribution. 
Section  \ref{experiments} presents results of our experiments for interpreting DNN models trained on one synthetic and two real-world data sets, using several state-of-the-art methods for causal effects estimation from observational data. 
Section \ref{discussion} concludes with a brief summary, a discussion of the related work, some caveats regarding the applicability of the proposed approach to interpretation of black box predictive models, and some promising directions for further research.

\section{Preliminaries} \label{prelim}


\subsection{Causal Effects}
The central problem in causal inference is determining
whether, and how, a change in a treatment \(T\) (e.g., surgery) leads to a change in some outcome \(Y\) (e.g., health status).

We will introduce the key notions when both the treatment and outcome are binary (for simplicity) before proceeding to consider the setting where the treatments are continuous-valued.

\subsection{Estimating Causal Effects of Discrete Treatments}

Let \(Y_i^{(t)}\) and \(Y_i^{(t^\prime)}\) be the {\em potential outcomes} when an individual \(i\), is exposed to treatments \(T = t\) and \(T = t^\prime\), respectively. The causal effect of $T$ on $Y$ is gauged with contrasting $Y_i^{(t)}$ and $Y_i^{(t^\prime)}$. An estimand of interest is the average causal effect, which is defined as follows if the treatment is binary:

\begin{definition} \label{def_ace} {\bf (Average Causal Effect (ACE))}
Consider a population of individuals each with the potential outcomes \(Y_i^{(t)}\) and \(Y_i^{(t^\prime)}\). The \emph{Average Causal Effect} of \(T\) on \(Y\) is defined as:
\begin{equation}
    ACE_{T}^{Y} = \mathbb{E}[Y_i^{(t)} - Y_i^{(t^\prime)}] = \mathbb{E}[Y_i^{(t)}] - \mathbb{E}[Y_i^{(t^\prime)}].
\end{equation}
\end{definition}

Because for each individual \(i\), the random variable \(T\) must either take the value $T=t$ or $T=t^\prime$ but not both, only one of the potential outcomes \(Y_i^{(t)}\) or \(Y_i^{(t^\prime)}\) is observable. For example, we observe the effect of a surgery on the individual's health status (e.g., cured), but cannot observe the effect of not having done surgery for the {\em same individual}. The observable outcome is called the factual outcome and the unobservable outcome is called the counterfactual outcome. 
Counterfactual inference requires us to estimate the outcome that {\em would have been observed}, had the treatment variable been assigned a value that is different from its observed value. Within the potential outcomes framework, the counterfactual outcomes are estimated from the observed outcome(s) for individual(s) of the opposite group that are {\em most similar} to the individual under consideration \cite{hernan2020causal}. This procedure may become unreliable in high-dimensional spaces \cite{aggarwal2001surprising}. To cope with this problem, we can use modern representation learning techniques \cite{bengio2013representation,tschannen2018recent} to map the data to a low-dimensional latent space before estimating the counterfactual outcomes \cite{schwab2018perfect, louizos2017causal,johansson2016learning}, or employ targeted maximum likelihood methods (TMLE)  \cite{van2011targeted} for counterfactual inference from observational data. 


\subsection{Causal Effects Under Continuous Treatment}
If the treatment is continuous, the potential outcomes can still be written as $Y_i^{(t)}$ where $t$ can take any value in a continuum, e.g., $t \in [t_{min},t_{max}]$ and one can contrast two values of the potential outcomes for any $t$ and $t^\prime$. The unit-level counterfactual outcomes are unobservable and an estimand of interest is $\mathbb{E}[Y_i^{(t_i)}]$, the average treatment effect. The potential outcomes framework \cite{rubin2005causal} offers a variety of estimators to estimate the average treatment effect in a continuous treatment regime, e.g., 
covariate balancing propensity score and its non-parametric counterpart  \cite{fong2018covariate}, propensity score weighting using generalized boosted models \cite{zhu2015boosting}, optimization based weighting \cite{zubizarreta2015stable}, inverse probability of treatment weighting in generalized linear models \cite{robins2000marginal}, propensity score weighting using the super learner \cite{van2007super,pirracchio2015improving} (see \cite{griefer2019weightit} for a review and Section \ref{estimators} for details). 

\subsection{Assumptions} \label{assumptions}
The potential outcomes approach to counterfactual inference \cite{hernan2020causal} makes a set of assumptions:
\begin{enumerate}
    \item 
Consistency: The potential outcome of any individual $i$ that has been exposed to a treatment $t$ is the realized outcome for that individual if the individual was treated with $t$. In other words, $T_i = t \implies Y^{(t)}_i = Y_i$; 
\item Positivity: In the case of discrete treatments, each possible treatment has non-zero probability; and in the case of continuous treatments, the conditional density of treatment given the covariates is non-negative for all covariates; 
\item Stable Unit Treatment Value Assumption (SUTVA): The potential outcomes of any individual do not depend on the treatment assigned to other individuals;  All treated (untreated or controlled) individuals receive the same version of treatment (control); 
\item Unconfoundedness: There are no unobserved confounders (i.e., variables that causally impact both the treatment and the outcome) and hence, adjusting for the observed confounders eliminates any bias in comparing the treated and untreated individuals -- an assumption  that is untestable from observational data alone. 
\end{enumerate}
Unconfoundedness and positivity (together referred to as strong ignorability) imply that the causal effect of the treatment is identifiable from observational data \cite{hernan2020causal}. In this paper, we will assume that the assumptions generally hold. See Section \ref{discussion} for discussion of how some of these assumptions can be relaxed if needed.

\section{Predictive Model Interpretation via Causal Attribution} \label{causal-attribution}

Let $D = \{{\bf X}_i,Y_i\}_{i=1}^{n}$ be a set of data samples from $\mathcal{X} \times \mathcal{Y}$, where $\mathcal{X}$ is the input space, or the domain of  feature vector ${\bf X}$, and $\mathcal Y$ is the domain of the outcome $Y$. Let $f_{\bf W}: \mathcal{X} \to \mathcal{Y}$, be a predictive model trained on $D$, so as to minimize some objective function, e.g., a suitable measure of error of the model's predictions $h({\bf X}_i)$, for ${\bf X}_i$ relative to the desired output $Y_i$, over the training data, using a suitable learning algorithm $L$. The resulting predictive model $f_{\bf W}$, given an input ${\bf X} \in \mathcal{X}$, produces an output $Y \in \mathcal{Y}$.

Recall that our goal is to interpret or explain a black box predictive model $f_{\bf W}$.  We aim to do so by attributing the responsibility for the model output to the model's inputs (i.e., features) \cite{sundararajan2017axiomatic, roscher2020explainable}. We focus on interpretation via causal attribution in the setting where the interpretation algorithm has no knowledge about the internal structure, or the parameters of the black box predictive model $f_{\bf W}$. Thus, if $f_{\bf W}$ is a deep neural network, the causal attribution algorithm can be expected to have no knowledge about the architecture of the  deep neural network, the objective function used to optimize its parameters, or the parameters of the trained network. 

We reduce the problem of interpreting the predictive model $h$ to the problem of determining the {\em causal effect} of each of the features of ${\bf X}$ on ${\bf Y}$ from observations of a sufficiently large data set of sample model inputs  and the corresponding model outputs. 
We aim to answer the following question: Why did $f_{\bf W}$ generate $Y=f_{\bf W}({\bf X})$ as the output for input ${\bf X}$? In other words, what is the causal effect of each of the features of ${\bf X}$ on $Y$? To answer this question,  for each  feature $X_j$ of ${\bf X}$, we designate $X_j$ as the ``treatment'' and estimate its average causal effect on $Y$. Consider the set of data samples  $\{{\bf V_i}, T_i,Y_i\}_{i=1}^n$ where ${\bf V_i} = {\bf X_i} \setminus T_i$. For each choice of $T$, we estimate the average causal effect of $T$ on $Y$, i.e., $ACE_T^Y$, yielding a vector of causal effects ${\bf \mu}=\{\mu_1,\ldots,\mu_m\}$ where each $\mu_j, j = 1,\ldots,m$ denotes the estimated causal effect of feature $X_j,j = 1,\ldots,m$ on $Y$.  We will use state-of-the-art methods for estimating the causal effect of each input on the output of the predictive model from  the observed input-output samples.

\begin{definition} {\bf (Causal Attribution)} \label{explanation_def} Given a predictive model $h: \mathcal{X} \to \mathcal{Y}$ trained on a data set $D$, causal attribution of the output $Y$ is ${\bf A} = \{a_1,\ldots,a_m\}$, where each $a_j =\mu_j$, is the causal effect of the feature $X_j$  on $Y$  iff $\mu_j$ is statistically significantly non-zero, and $a_j=0$ otherwise. 
\end{definition}
The non-zero causal effects included in the causal attribution are identified by testing the null hypothesis $H_0:ACE_{x_j}^{y} = 0$, $\forall j \in \{1,\ldots, m\}$, where $x_j$ denotes the model inputs and $y$ the model output. If $H_0$ is rejected at a chosen level of statistical significance, then $x_j$ is causally responsible for output $y$ and the degree of its influence is gauged by the estimated causal effect.

\subsection{Estimating Causal Effects} \label{estimators}

We use potential outcomes based state-of-the-art methods for estimating causal effects of continuous treatments on outcomes \cite{griefer2019weightit}.
The mechanism of the methods in this paper (described in detail below) involves two steps: 
\begin{enumerate}
    \item Estimation of  weights, $w_i = f(T_i,{\bf X_i},\epsilon_i)$, for each data sample to ideally achieve independence between their features and the treatment and thus, deconfounding (or balancing) the treatment and outcome;
    \item Modeling the outcome $Y_i = g(T_i,{\bf X_i},\epsilon_i)$, as a function of the {\em weighted} data samples.
\end{enumerate}
If either the treatment model $f(\cdot)$ or the outcome model $g(\cdot)$ are correctly specified, the estimated result is an unbiased estimate of the average causal effect of the treatment on the outcome. The methods that we use have been shown to be effective in a variety of causal effect estimation applications \cite{chen2016personalized,kreif2019machine,zhao2019causal,khademi2019fairness,fong2018covariate,barlow2020liberal} and are listed as follows:
\begin{enumerate}

    \item {\bf Covariate Balancing Propensity Score (CBPS)}  \cite{fong2018covariate} which is an estimate of the generalized propensity score (GPS) or the conditional propensity density, defined as $r(t,x) = f(T_i=t_i \, | \, {\bf X}_i = {\bf x}_i)$, while achieving covariate balance. Generalized propensity score is a generalization, to continuous treatment regimes \cite{hirano2005propensity,imai2004causal}, of the propensity score \cite{rosenbaum1983central}, a well-established and very commonly used distance measure for counterfactual estimation in binary treatment regimes, which is defined as $Pr(T_i = 1 \, | \, {\bf X}_i = {\bf x}_i)$. CBPS estimates the GPS under covariate balancing constraints to maximize covariate balance, thereby avoiding the need for an iterative  model fitting process until the desired balance is achieved. 
    \item {\bf Non-Parametric Covariate Balancing Propensity Score (NPCBPS)} \cite{fong2018covariate} which is a non-parametric approach that maximizes the likelihood of the observed data under covariate balancing constraints for computing the GPS weights.
    \item {\bf Propensity Score Weighting Using Generalized Boosted Models (PSWGBM)} \cite{zhu2015boosting} which estimates the propensity scores using generalized boosted models. In this method, we use the mean of the Spearman correlation statistic to maximize balance between data points of different treatments.
    \item {\bf Optimization-Based Weighting (OPTWEIGHT)} \cite{zubizarreta2015stable} which solves a  quadratic optimization problem constrained to achieve covariate balance (we used a million iterations and set the absolute difference in means of the weighted features  $\delta \leq 0.1$ as the convergence criterion).
    \item {\bf Inverse Probability of Treatment Weighting in Generalized Linear Models (IPTW)} \cite{robins2000marginal} which estimates the propensity scores using generalized linear models and weights each data sample according to the inverse of the estimated propensity score. 
    \item {\bf Propensity Score Weighting Using Super Learner (SUPER)} \cite{van2007super,pirracchio2015improving} which is designed to be robust to misspecification of the treatment model, the outcome model, or both. It estimates the propensity score based on an optimized weighted combination of an ensemble of candidate prediction models optimized using cross validation.
\end{enumerate}

\section{Experiments and Results} \label{experiments}
We proceed to report results of our experiments with the proposed methods for interpretation via causal attribution of black box predictive models. To facilitate direct comparisons with \cite{chattopadhyay2019neural}, we used the trained neural network as the model to be interpreted, and the inputs and outputs of neural network as observational data  for estimating the causal effect of each input of the neural network on its output. We also report results on an additional real-world data set and the trained DNN from \cite{ancona2019explaining}. 

\subsection{Data Sets and DNN Models Used}

\subsubsection{{\bf Synthetic data set}} We used a synthetic data set from \cite{chattopadhyay2019neural},  generated following \cite{hochreiter1997long} where a data sample consists  of 12 features, $f_1, \ldots, f_{12}$, sampled from the normal distribution according to the following procedure: $\forall i \in \{4, \ldots, 12\}: f_i \sim \mathcal{N}(0,0.2)$, and 
\begin{enumerate}
    \item either (with probability 0.5) $\forall j \in \{1,2,3\}$: $f_j \sim \mathcal{N}(1,0.2)$ and set the class label to 1,
    \item or (with probability 0.5) $\forall j \in \{1,2,3\}$: $f_j \sim \mathcal{N}(-1,0.2)$ and set the class label to 0.
\end{enumerate}

The data generation procedure ensures that only the first 3 features are responsible for class labels. We used the 3-layer neural network used in the experiments of  \cite{chattopadhyay2019neural}. We generated a 1000 data samples from the synthetic data generator described above and obtained the predicted outputs of the neural network on each of the data samples. We used the resulting data for causal attribution.

\subsubsection{{\bf MNIST data set}} 
We used a convolutional neural network (CNN) trained on the MNIST training data to classify the handwritten digits of $0,\ldots,9$. We ran each  test image with 784 features ($28\times28$ pixels) through the trained CNN  and recorded the predicted class labels. For ease of interpretation, we trained an Auto-Encoder (AE) on the test data to reduce the dimensions from 784 to 10 latent variables, $Z_0,\ldots,Z_9$, and labeled each 10-dimensional feature vector with the corresponding CNN-predicted class label for the $28\times28$ or 784 pixel image. We used the resulting data for causal attribution of the CNN output to the 10 latent variables. \footnote{Note that we could have performed the same analysis with 784 pixel variables, but we chose the 10 latent variables setup used here for ease of visualization, and to demonstrate the power of the proposed method to perform causal attribution with respect to {\em any} variables that are derived from the input variables (which in this case are the 10 latent variables).}

\subsubsection{{\bf Parkinson's Disease (PD) Telemonitoring data set}}  The PD telemonitoring data set \cite{tsanas2009accurate} provides age, gender, and 16 (at-home) biomedical voice measurements (processed from voice recordings) obtained from PD patients (see \cite{tsanas2009accurate} for details) along with their scores on the Universal Parkinson's Disease Rating Scale (UPDRS) which measures the severity and progression of the Parkinson's disease. We used the DNN trained by  \cite{ancona2019explaining} on the PD data and used it to predict UPDRS for each PD data sample. We used the inputs and outputs of the DNN for each of the PD data samples for causal attribution.


\subsection{Causal Attribution of Black Box Models}
In keeping with our goal of interpreting black box models, we used {\em only} the observed inputs and outputs of the model (and not any information related to the structure, parameters, or the algorithm used to optimize the model parameters) for causal attribution of the model output relative to the model inputs. To estimate the causal effects of the model inputs on the model output, we used the R package WeightIt \cite{griefer2019weightit} (version 0.7.1). Unless otherwise noted, we used the default parameter settings for each method. In identifying non-zero causal effects, we set the significance level $\alpha = 0.05$.  

\subsection{From Causal Attributions to Contrastive Explanations}
We also explored how the causal attributions of model output with respect to model inputs can be used to explain why the model output for a specific input data pattern differs from that for another. Specifically, we focus on the features that have large causal attributions for the model prediction, and examine how the two data samples being contrasted differ with respect to the values of those features.

\subsection{Experimental Results}
\subsubsection{{\bf Synthetic data set}}
We estimated the causal effects of all features of the data set on the class label predicted by the neural network and report the results in Table \ref{table:synthetic}. 

\begin{table*}[t]
\caption{\label{table:synthetic} Estimates and p-value significance of causal effects of input features on the outputs of the neural network trained using the synthetic data set.}
\resizebox{\textwidth}{!}{\begin{tabular}{lcccccccccccc}
\toprule
        & \multicolumn{2}{c}{CBPS}   & \multicolumn{2}{c}{NPCBPS} & \multicolumn{2}{c}{PSWGBM}    & \multicolumn{2}{c}{IPTW}   & \multicolumn{2}{c}{OPTWEIGHT} & \multicolumn{2}{c}{SUPER}  \\ \midrule
Feature & Est. & P         & Est. & P         & Est. & P         & Est. & P         & Est.   & P         & Est. & P         \\ \midrule
f1      & 1.69     & \textless{}0.01 & 1.36     & \textless{}0.01 & 0.62     & \textless{}0.01 & 1.81     & \textless{}0.01 & 2.53       & \textless{}0.01  & 1.53     & \textless{}0.01 \\
f2      & 1.58     & \textless{}0.01 & 1.20     & \textless{}0.01 & 1.51     & \textless{}0.01 & 1.60     & \textless{}0.01 & 2.67       & \textless{}0.01  & 1.24     & \textless{}0.01 \\
f3      & 1.83     & \textless{}0.01 & 1.00     & \textless{}0.01 & 0.54     & \textless{}0.01 & 1.93     & \textless{}0.01 & 2.26       & \textless{}0.01  & 1.76     & \textless{}0.01 \\
f4      & 0.00     & 0.96            & 0.00     & 0.96            & 0.04     & 0.63            & -0.01    & 0.94            & 0.00       & 0.97             & 0.02     & 0.80            \\
f5      & 0.07     & 0.42            & 0.07     & 0.43            & 0.09     & 0.33            & 0.07     & 0.43            & 0.07       & 0.41             & 0.07     & 0.44            \\
f6      & -0.11    & 0.21            & -0.11    & 0.20            & -0.04    & 0.67            & -0.11    & 0.19            & -0.11      & 0.21             & -0.07    & 0.39            \\
f7      & 0.16     & 0.06            & 0.16     & 0.05            & 0.17     & 0.05            & 0.16     & 0.05            & 0.16       & 0.06             & 0.14     & 0.09            \\
f8      & -0.04    & 0.67            & -0.04    & 0.66            & -0.03    & 0.76            & -0.03    & 0.71            & -0.03      & 0.69             & -0.05    & 0.54            \\
f9      & -0.04    & 0.67            & -0.04    & 0.67            & 0.03     & 0.69            & -0.04    & 0.68            & -0.03      & 0.70             & -0.02    & 0.83            \\
f10     & 0.15     & 0.09            & 0.15     & 0.08            & 0.16     & 0.06            & 0.15     & 0.08            & 0.15       & 0.08             & 0.15     & 0.08            \\
f11     & 0.07     & 0.42            & 0.08     & 0.39            & 0.02     & 0.81            & 0.07     & 0.44            & 0.07       & 0.41             & 0.06     & 0.52            \\
f12     & -0.19    & 0.04            & -0.19    & 0.04            & -0.14    & 0.11            & -0.19    & 0.04            & -0.19      & 0.04             & -0.12    & 0.18            \\ \bottomrule
\end{tabular}}
\end{table*}

Upon testing the null hypothesis (that the causal effect of each feature is 0), we find, from all of the causal effect estimation methods, that the output of the neural network is causally impacted by only the features $f_1,f_2,f_3$ and not  $f_4,\ldots,f_{12}$. This finding is in agreement with the design of the simulated data generator (which ensures that only the features $f_1,f_2,f_3$ determine the class label) and the results reported in \cite{chattopadhyay2019neural,sundararajan2017axiomatic}.

\noindent{\bf Example:} Consider, for example, the neural network constructed from the simulated data where we found the features $\{f_1,f_2,f_3\}$ to have large causal effects on the model's predictions. Now suppose a user wants to understand why the model outputs class label $0$ for sample
\begin{itemize}
    \item ${\bf S}_1$ = [-0.89, -1.11, -0.85, -0.33, -0.47, 0.23, -0.20, 0.13, -0.17, 0.35, -0.22, 0.04], 
\end{itemize}
and class label 1 for sample 
\begin{itemize}
    \item ${\bf S}_2$ = [0.81, 0.95, 1.11, 0.01, 0.12, -0.22, 0.23, 0.18, 0.10, 0.18, -0.14, -0.02].
\end{itemize} The answer to the user's question can be obtained by (i) noting that only the first three features, $\{f_1,f_2,f_3\}$, have non-zero causal attributions and (2) recognizing that there is a clear shift in the values of the features $\{f_1,f_2,f_3\}$ from the mean of the features for class $0$ samples towards the mean for class $1$ samples. 

\subsubsection{{\bf MNIST data set}} Using all causal effect estimation methods described in Section \ref{estimators}, we estimated the causal effect of latent variables (of test data) $Z_0,\ldots,Z_9$, obtained using the AE, on the class label $C_k: \forall \, k \in \{0,\ldots,9\}$, predicted by the CNN. We obtained causal attributions of the predicted class wih respect to the latent variables. To validate our results, we proceeded with two steps: (1) We intervened on each of the causal latent variables by manually setting their value to zero and reconstructed the image. If a latent variable has a large non-zero causal attribution for the predicted  label, we expect the reconstructed image to deviate from its original appearance; and (2) We intervened on all of the non-causal latent variables, i.e., those with low causal attributions, by setting their value to zero and reconstructed the image using only the remaining variables. We expect the resulting reconstruction to be similar to the original image.

\begin{figure}[!t]
    \centering
    \subfloat{\includegraphics[width=0.115\textwidth,trim={1.5cm 0 1.5cm 0},clip]{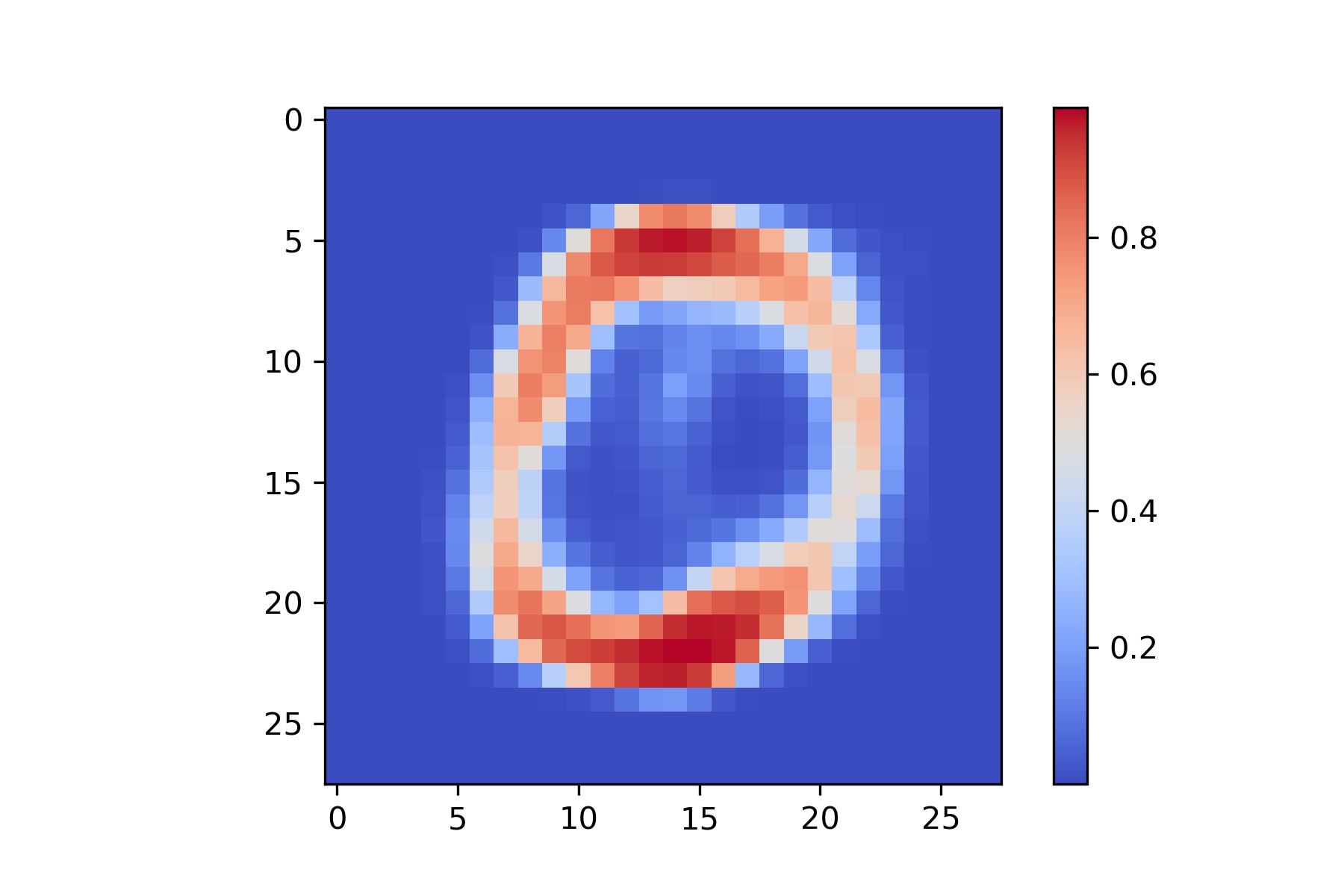}}
    \subfloat{\includegraphics[width=0.115\textwidth,trim={1.5cm 0 1.5cm 0},clip]{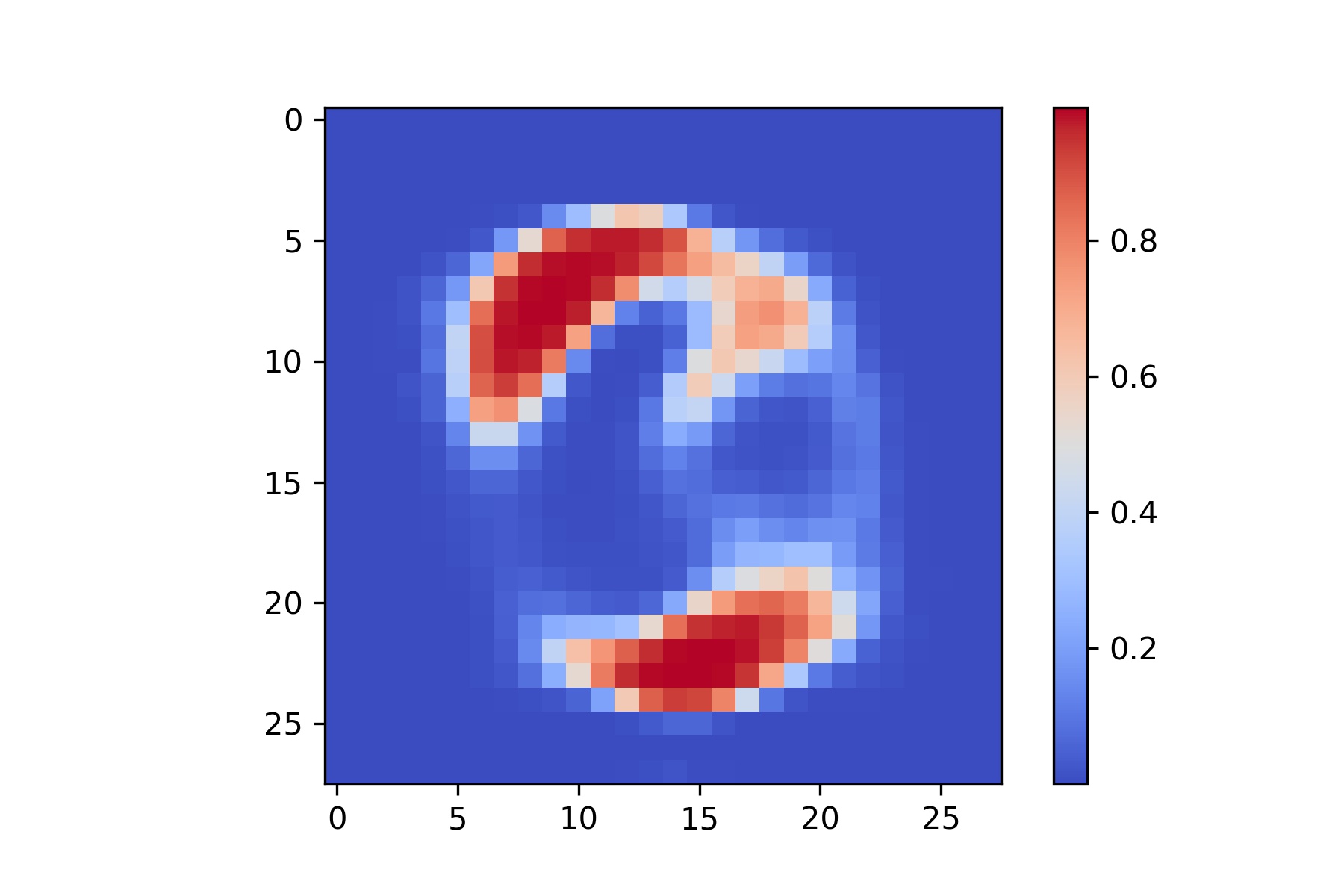}}
    \subfloat{\includegraphics[width=0.115\textwidth,trim={1.5cm 0 1.5cm 0},clip]{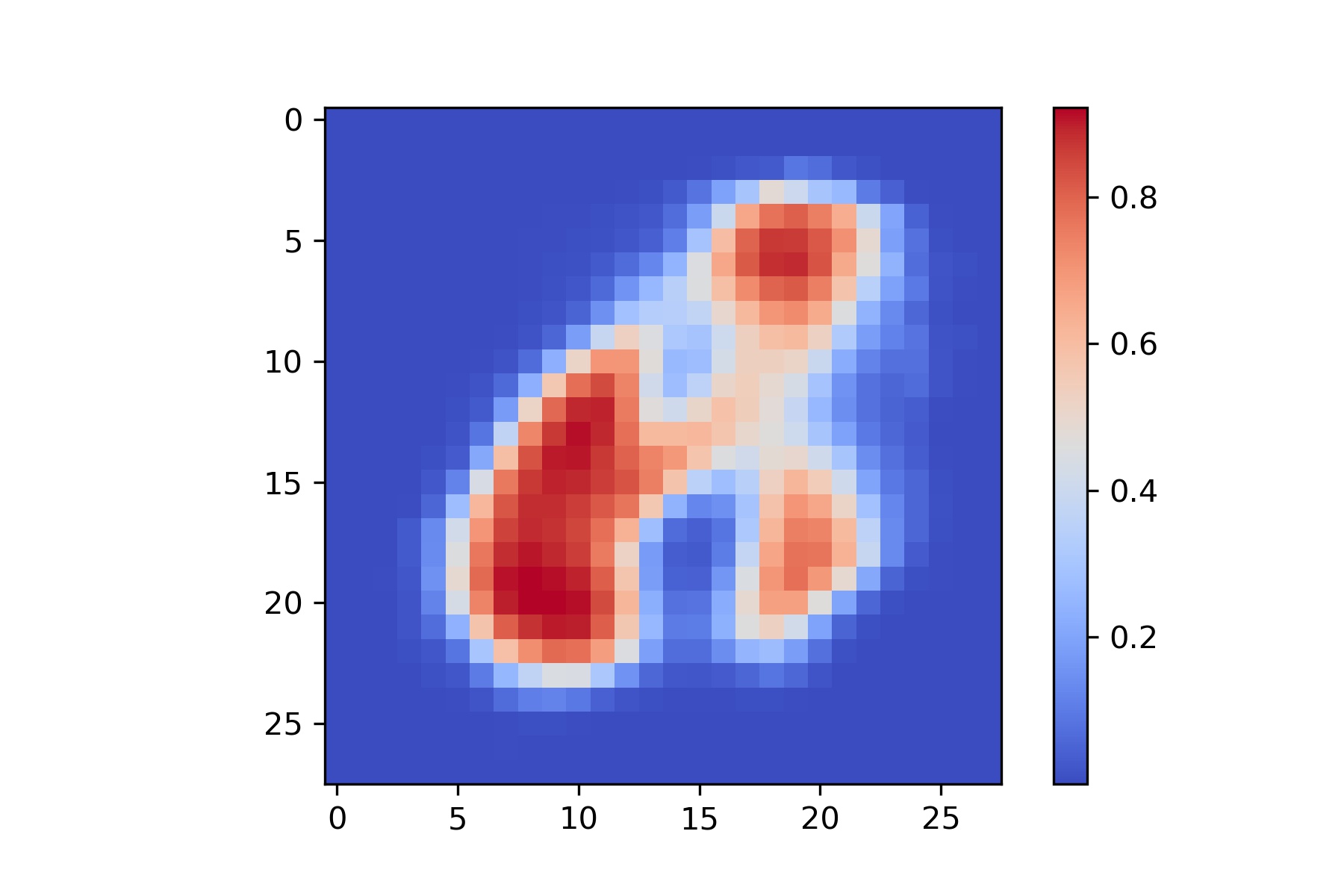}}
    \subfloat{\includegraphics[width=0.115\textwidth,trim={1.5cm 0 1.5cm 0},clip]{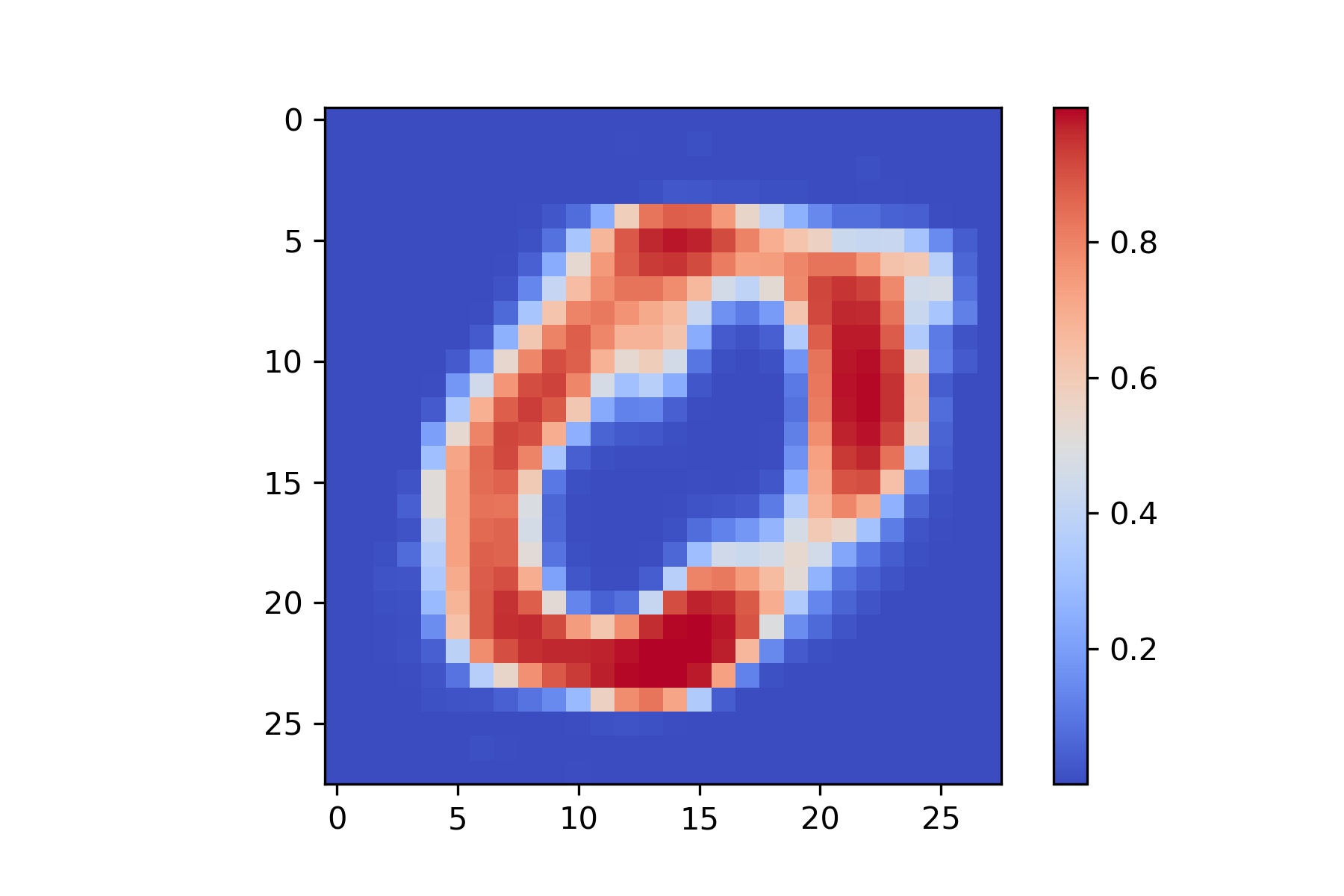}}
    \\
    \subfloat{\includegraphics[width=0.115\textwidth,trim={1.5cm 0 1.5cm 0},clip]{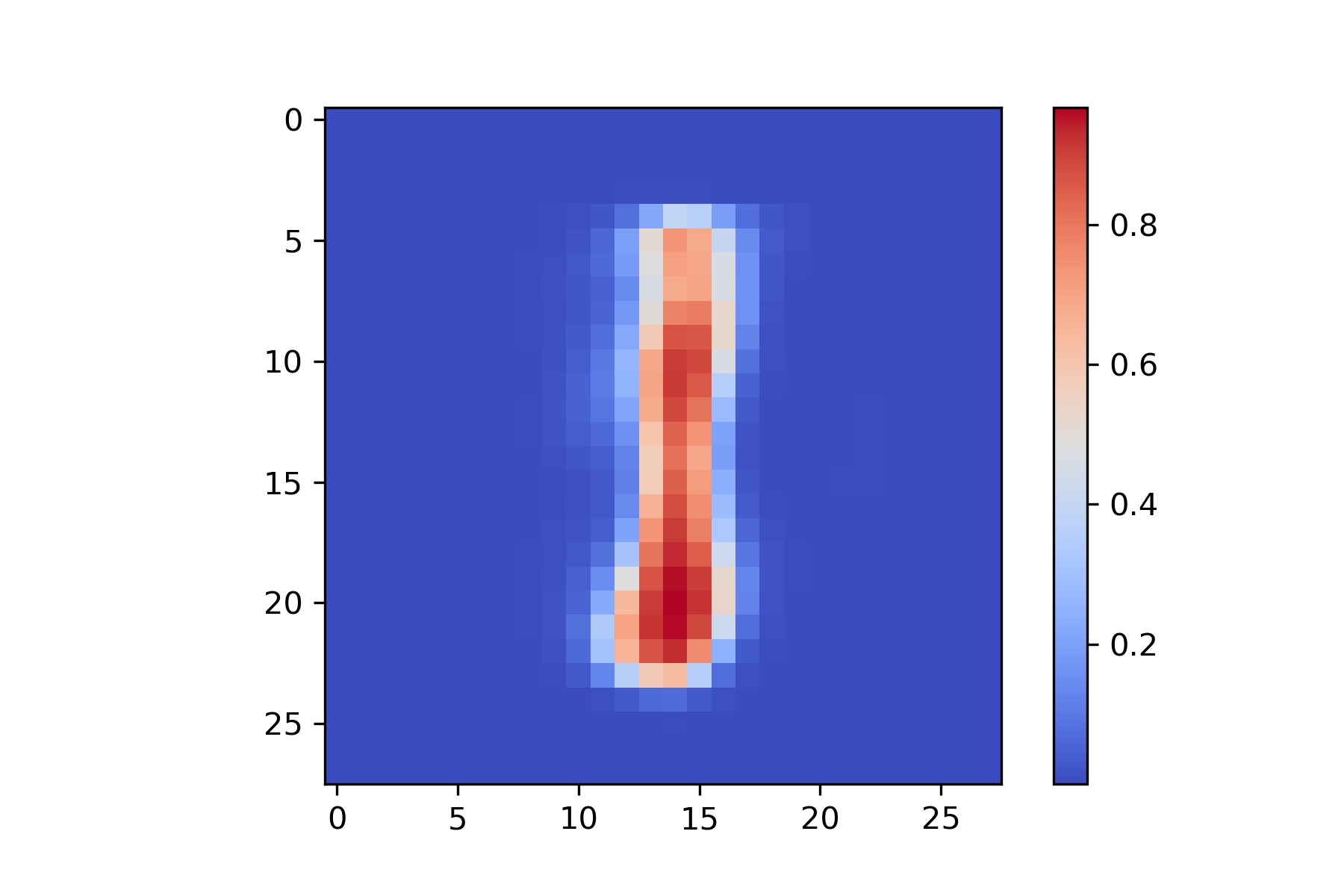}}
    \subfloat{\includegraphics[width=0.115\textwidth,trim={1.5cm 0 1.5cm 0},clip]{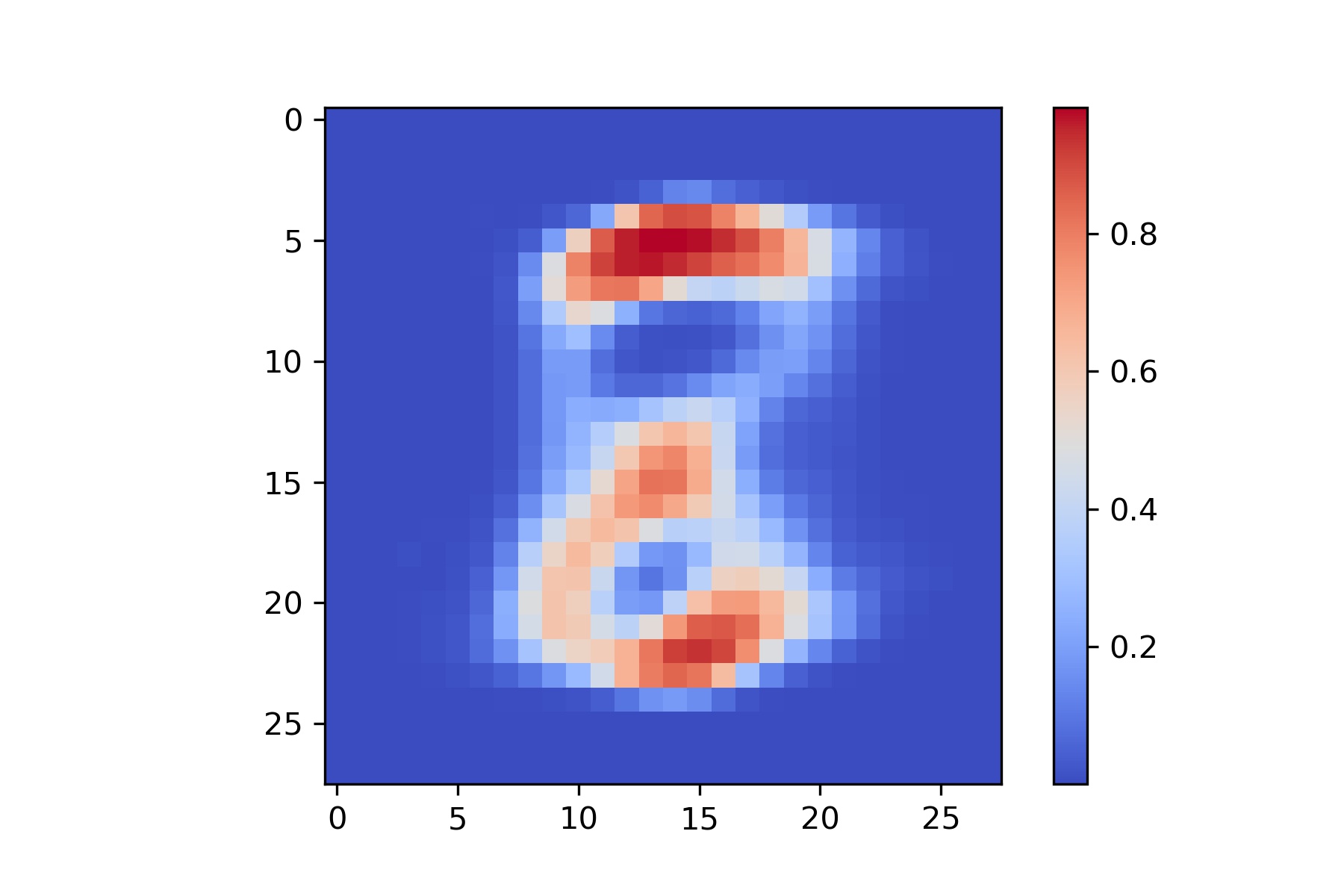}}
    \subfloat{\includegraphics[width=0.115\textwidth,trim={1.5cm 0 1.5cm 0},clip]{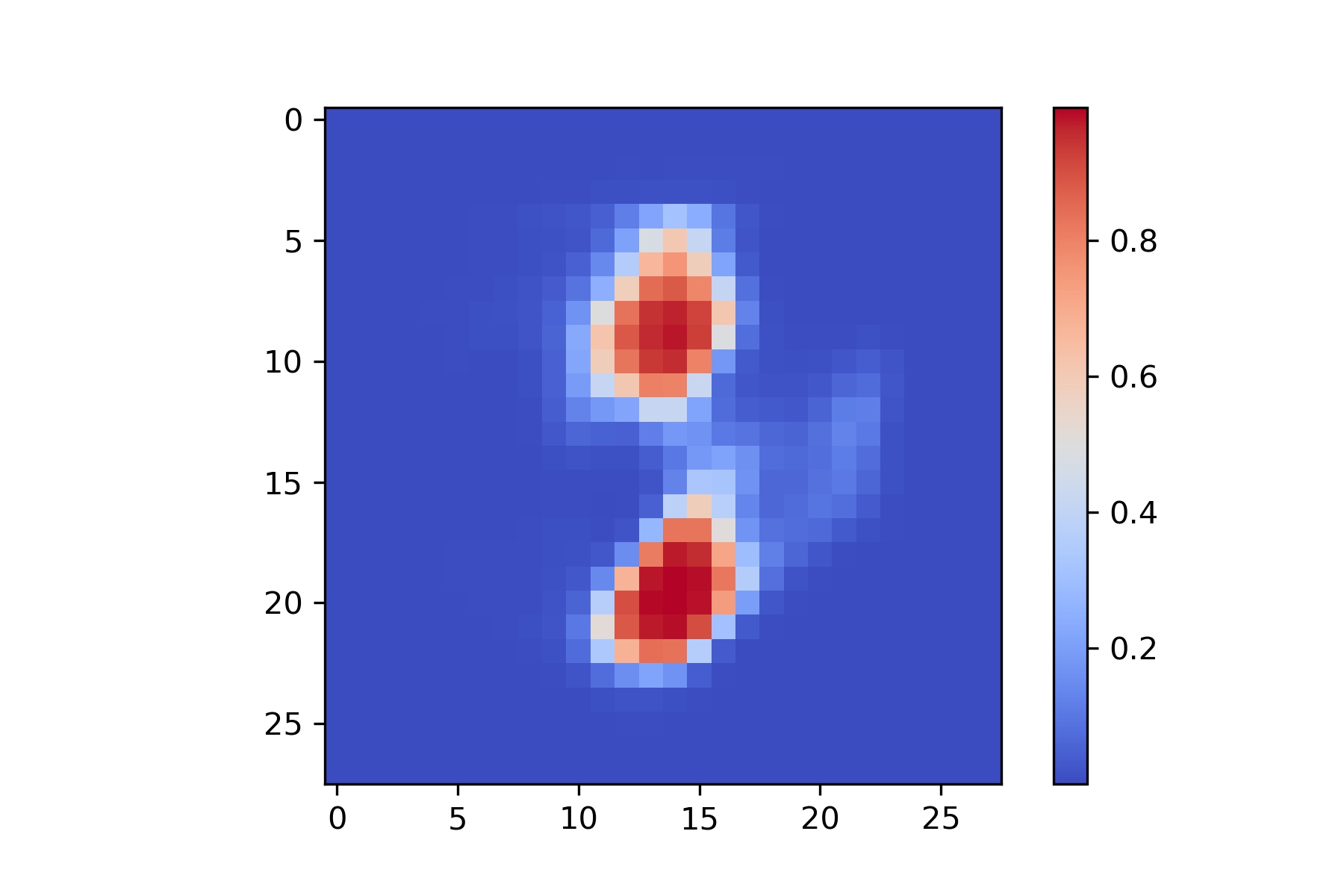}}
    \subfloat{\includegraphics[width=0.115\textwidth,trim={1.5cm 0 1.5cm 0},clip]{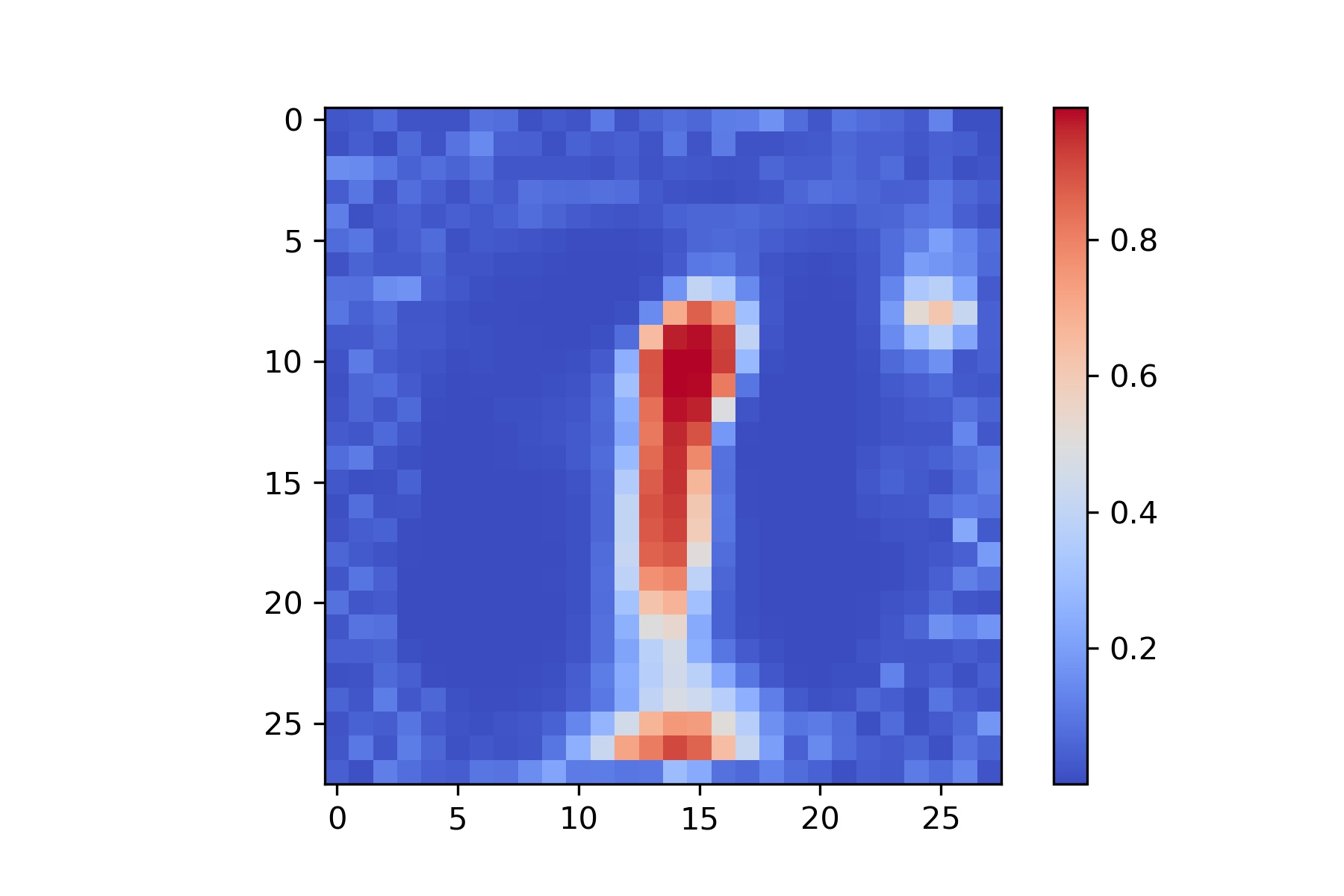}}
    \\
    \subfloat{\includegraphics[width=0.115\textwidth,trim={1.5cm 0 1.5cm 0},clip]{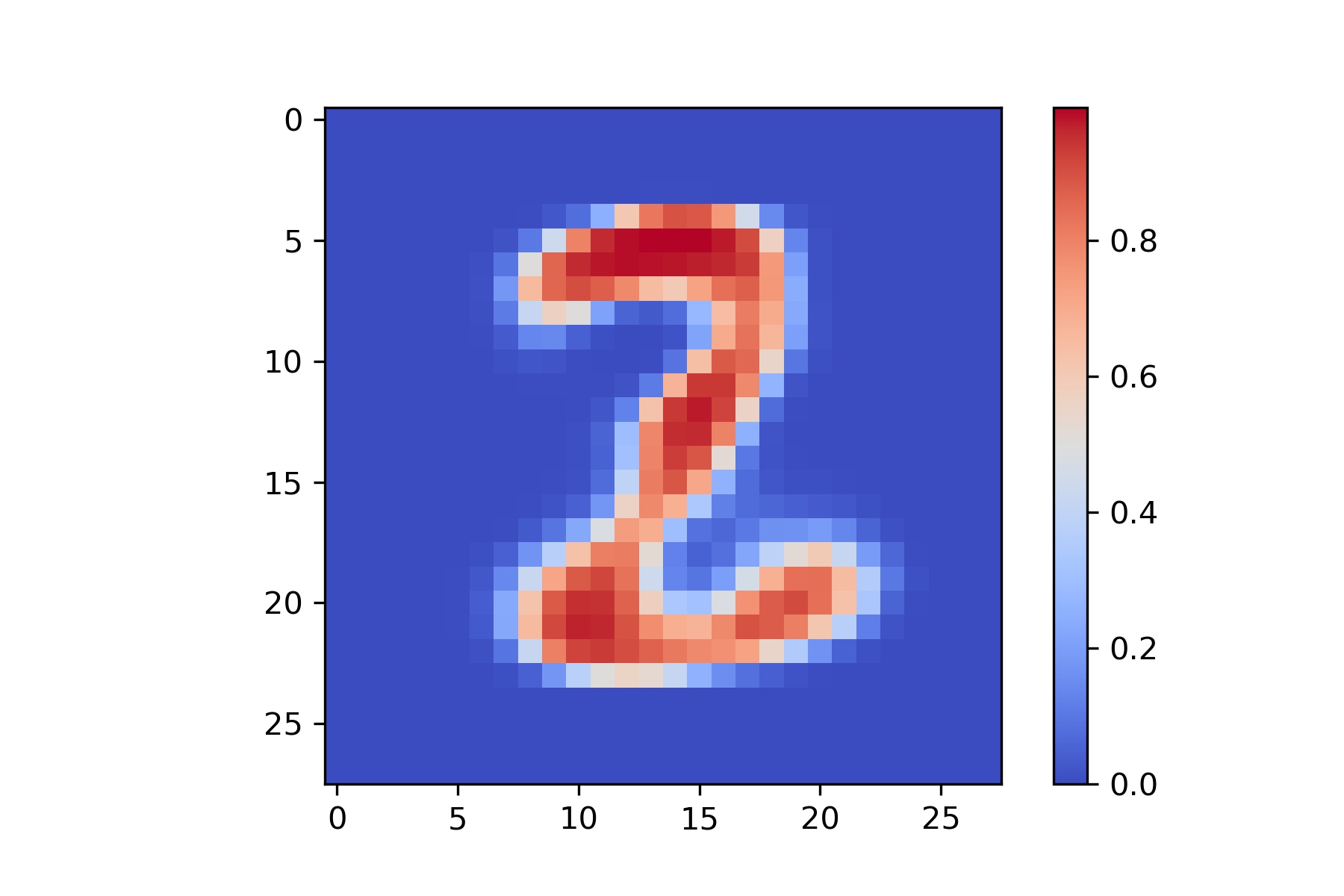}}
    \subfloat{\includegraphics[width=0.115\textwidth,trim={1.5cm 0 1.5cm 0},clip]{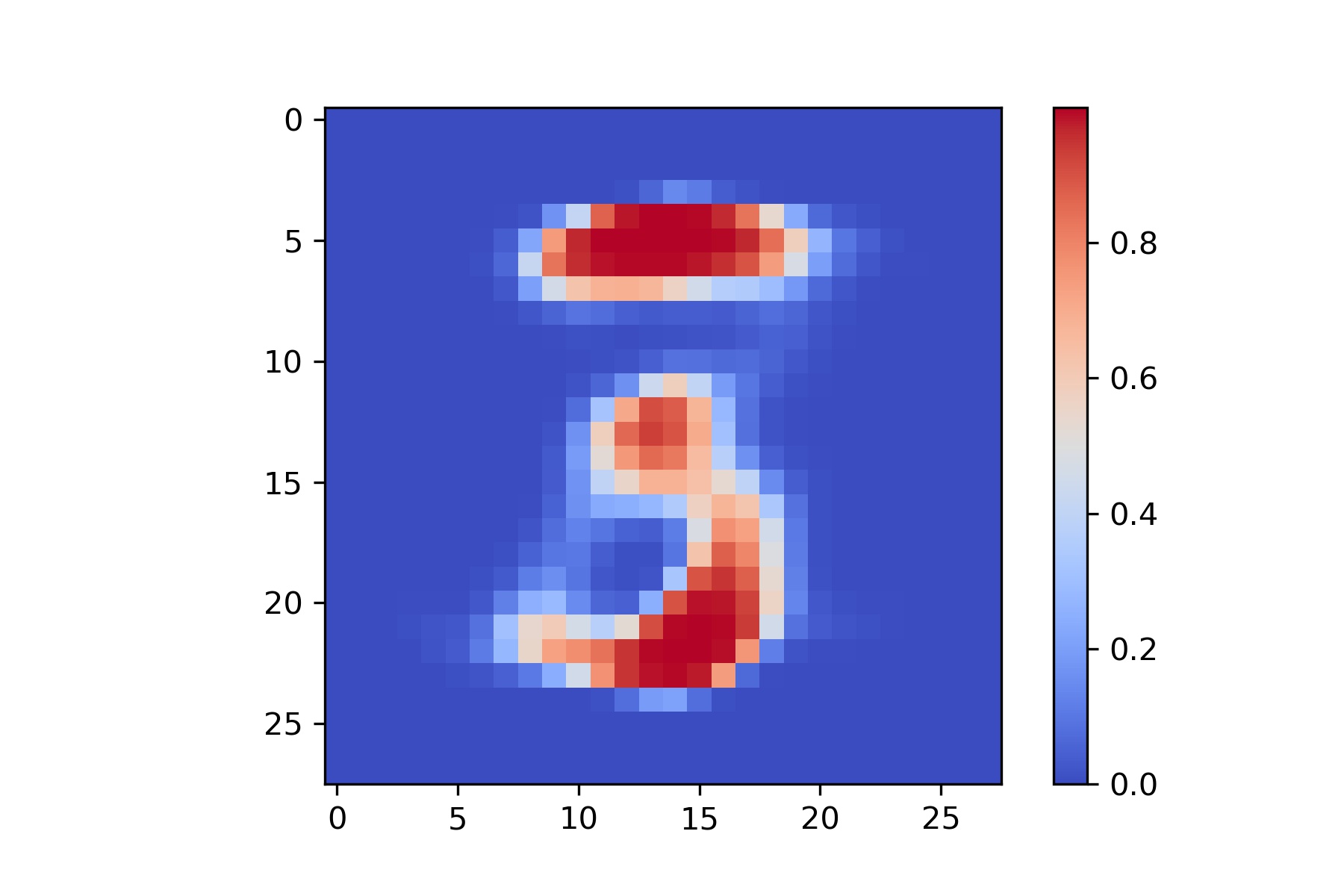}}
    \subfloat{\includegraphics[width=0.115\textwidth,trim={1.5cm 0 1.5cm 0},clip]{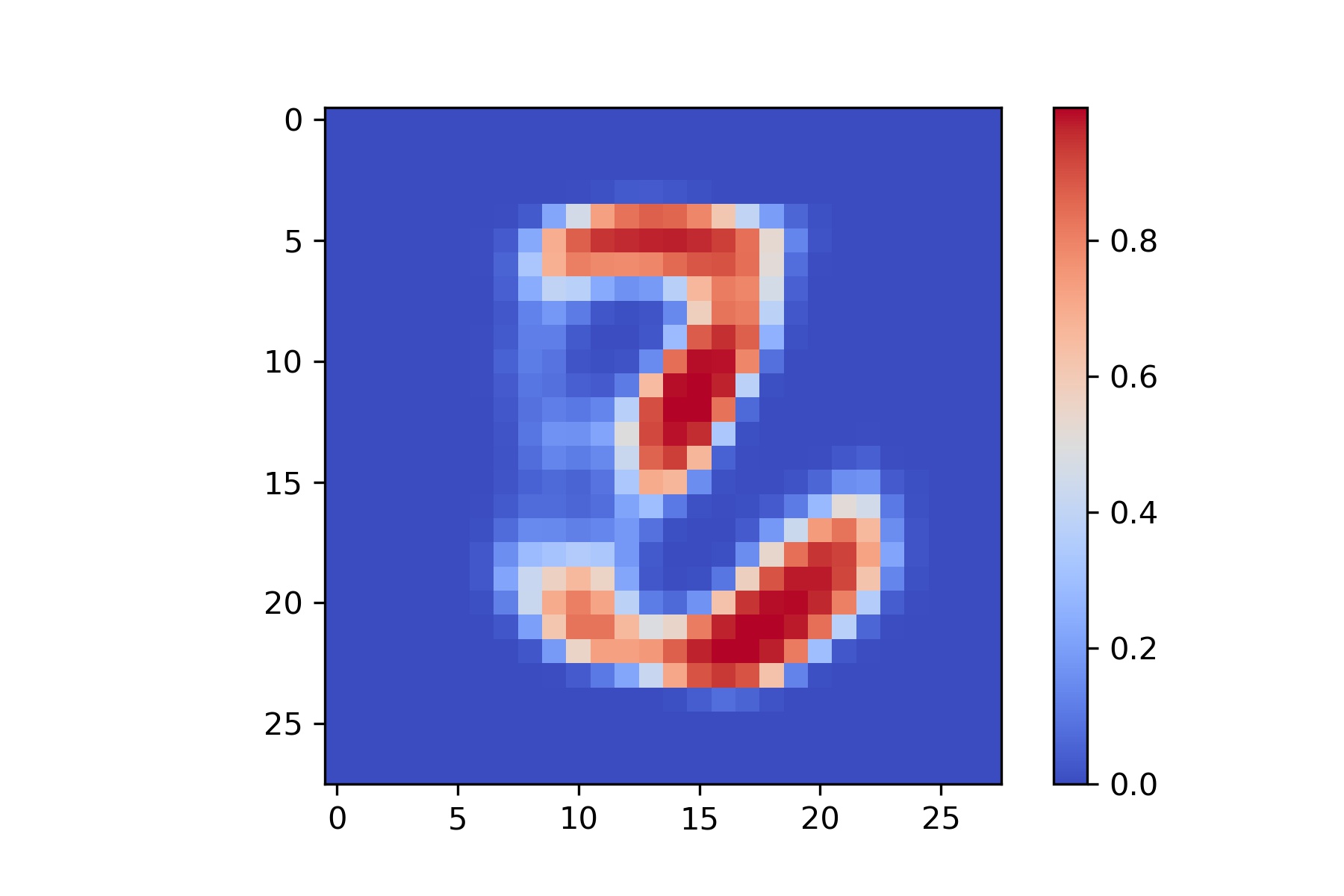}}
    \subfloat{\includegraphics[width=0.115\textwidth,trim={1.5cm 0 1.5cm 0},clip]{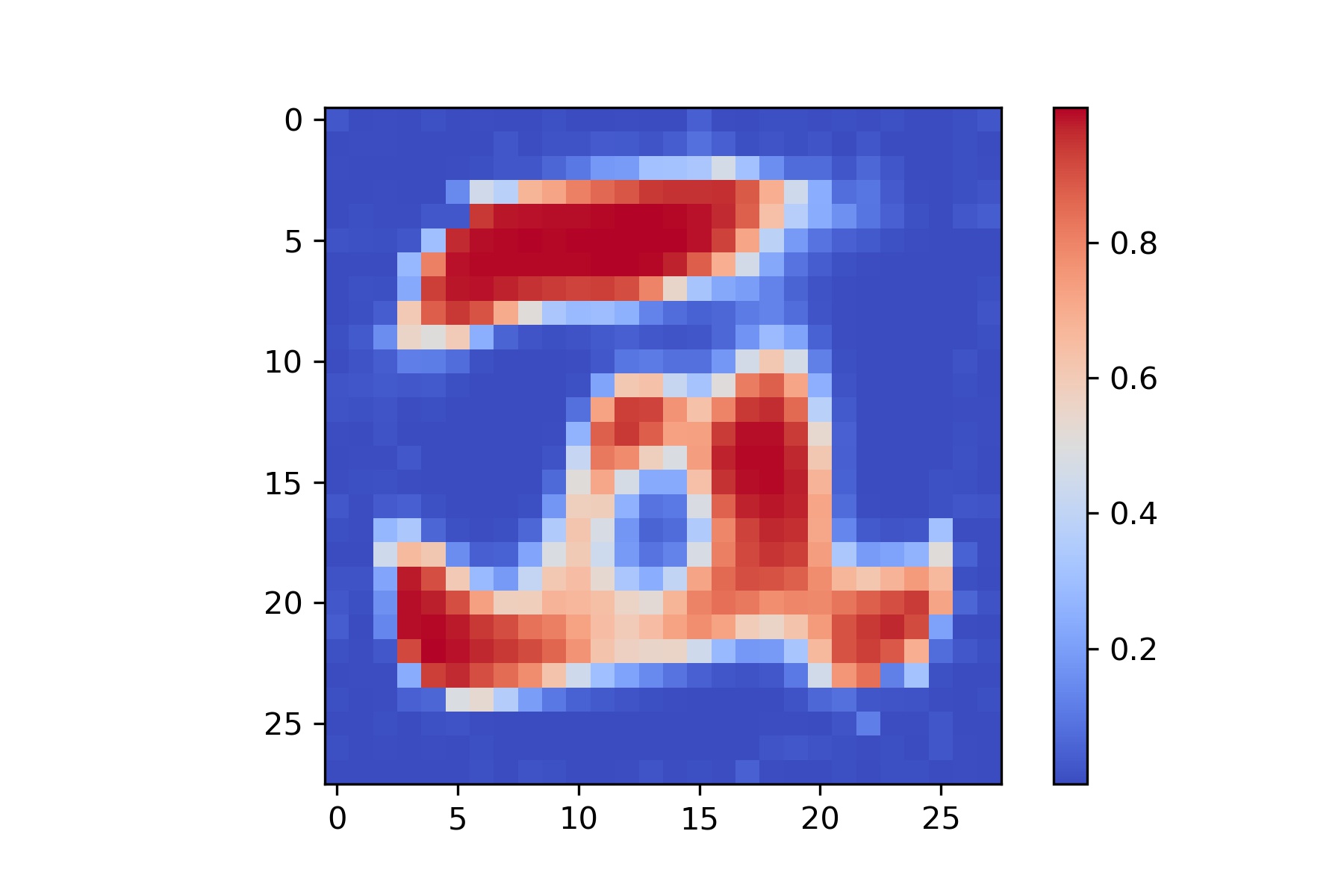}}
    \\    
    
    \subfloat{\includegraphics[width=0.115\textwidth,trim={1.5cm 0 1.5cm 0},clip]{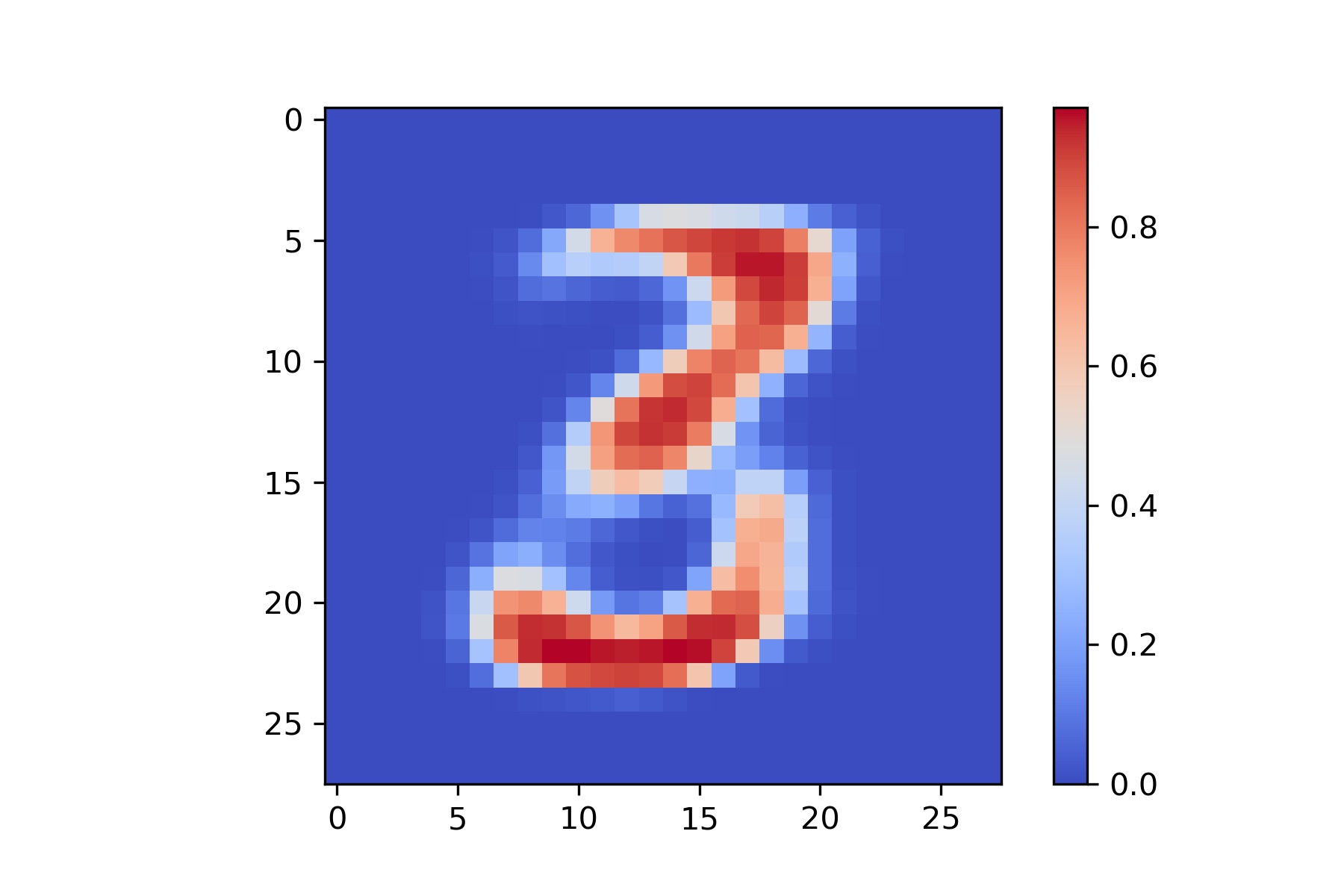}}
    \subfloat{\includegraphics[width=0.115\textwidth,trim={1.5cm 0 1.5cm 0},clip]{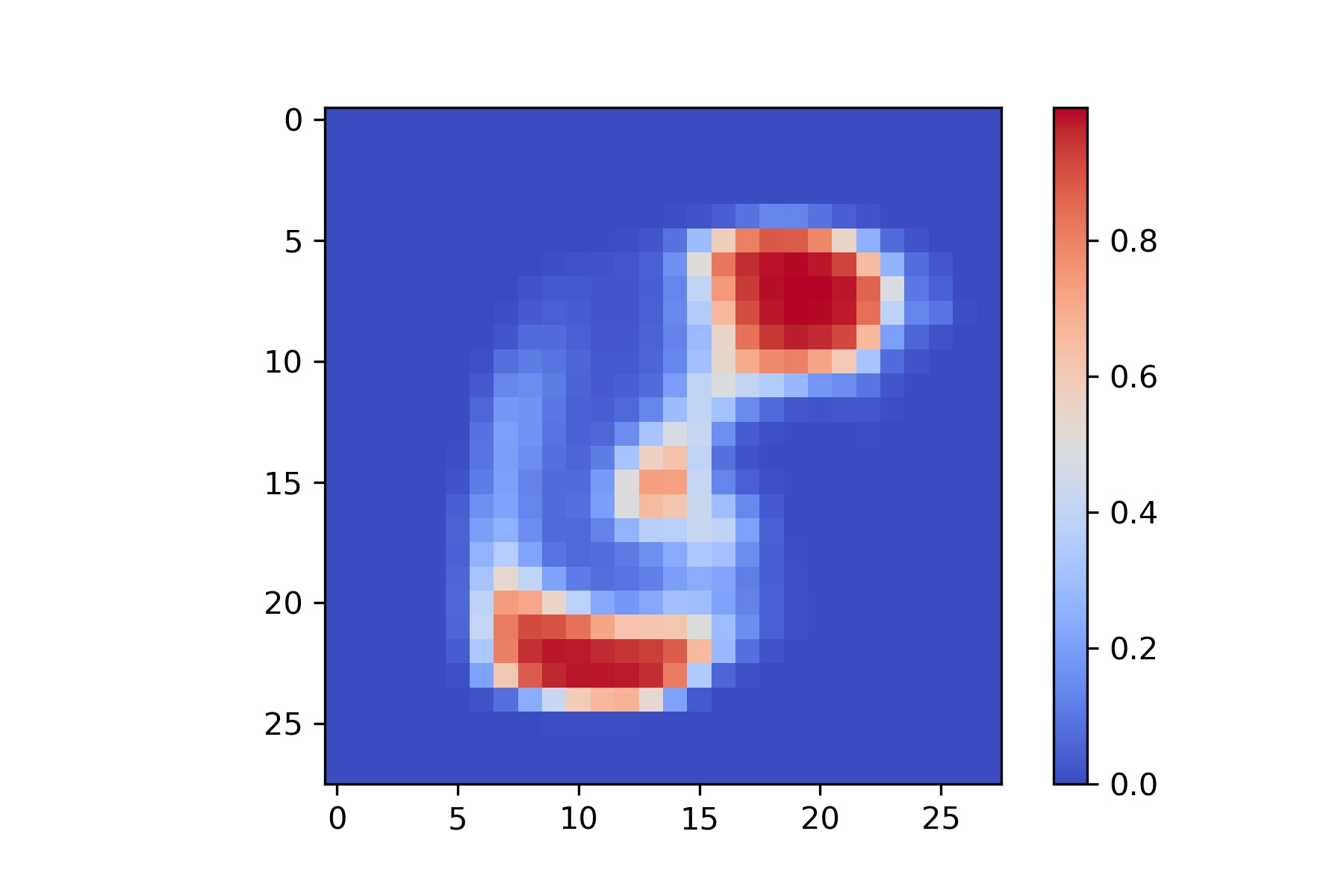}}
    \subfloat{\includegraphics[width=0.115\textwidth,trim={1.5cm 0 1.5cm 0},clip]{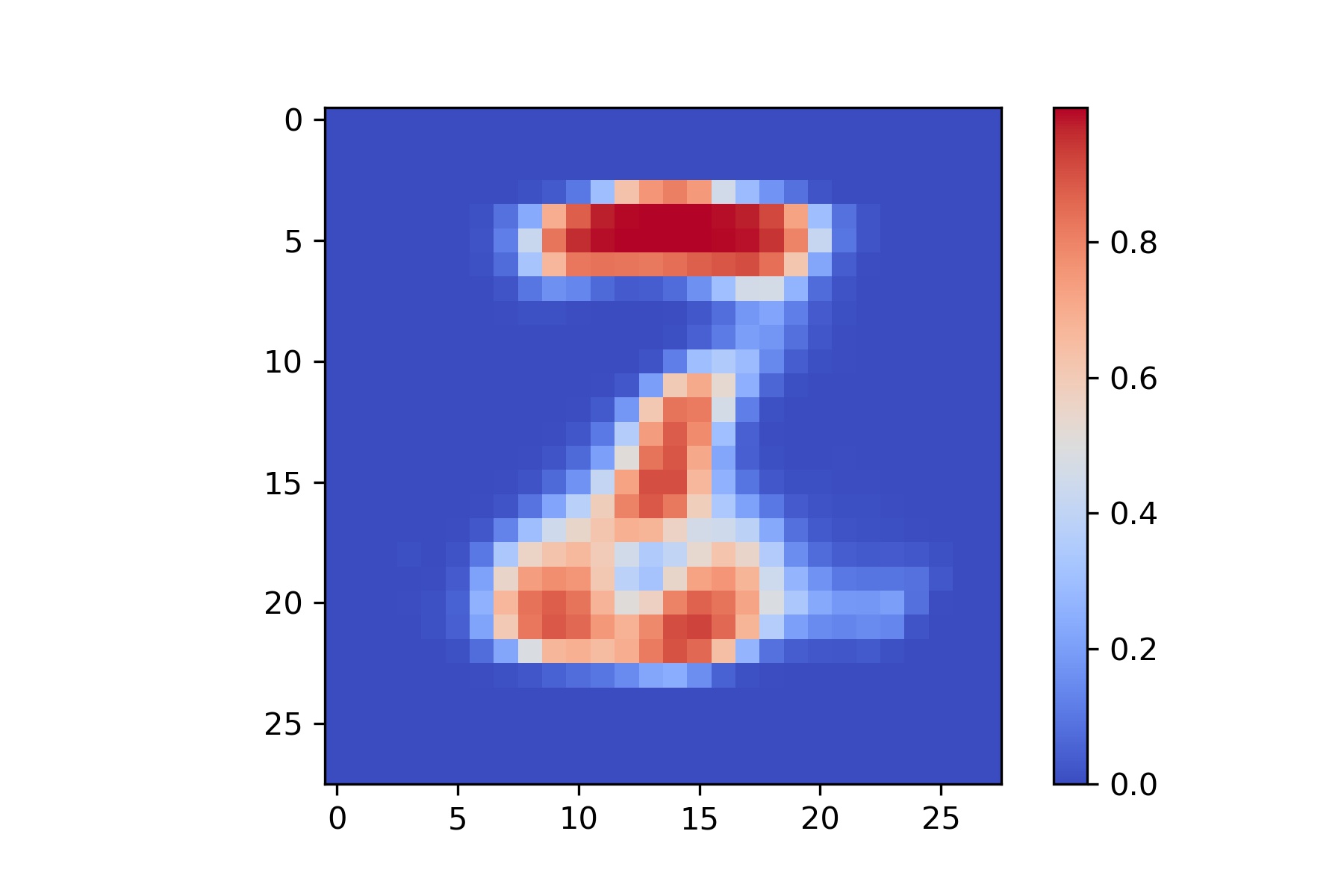}}
    \subfloat{\includegraphics[width=0.115\textwidth,trim={1.5cm 0 1.5cm 0},clip]{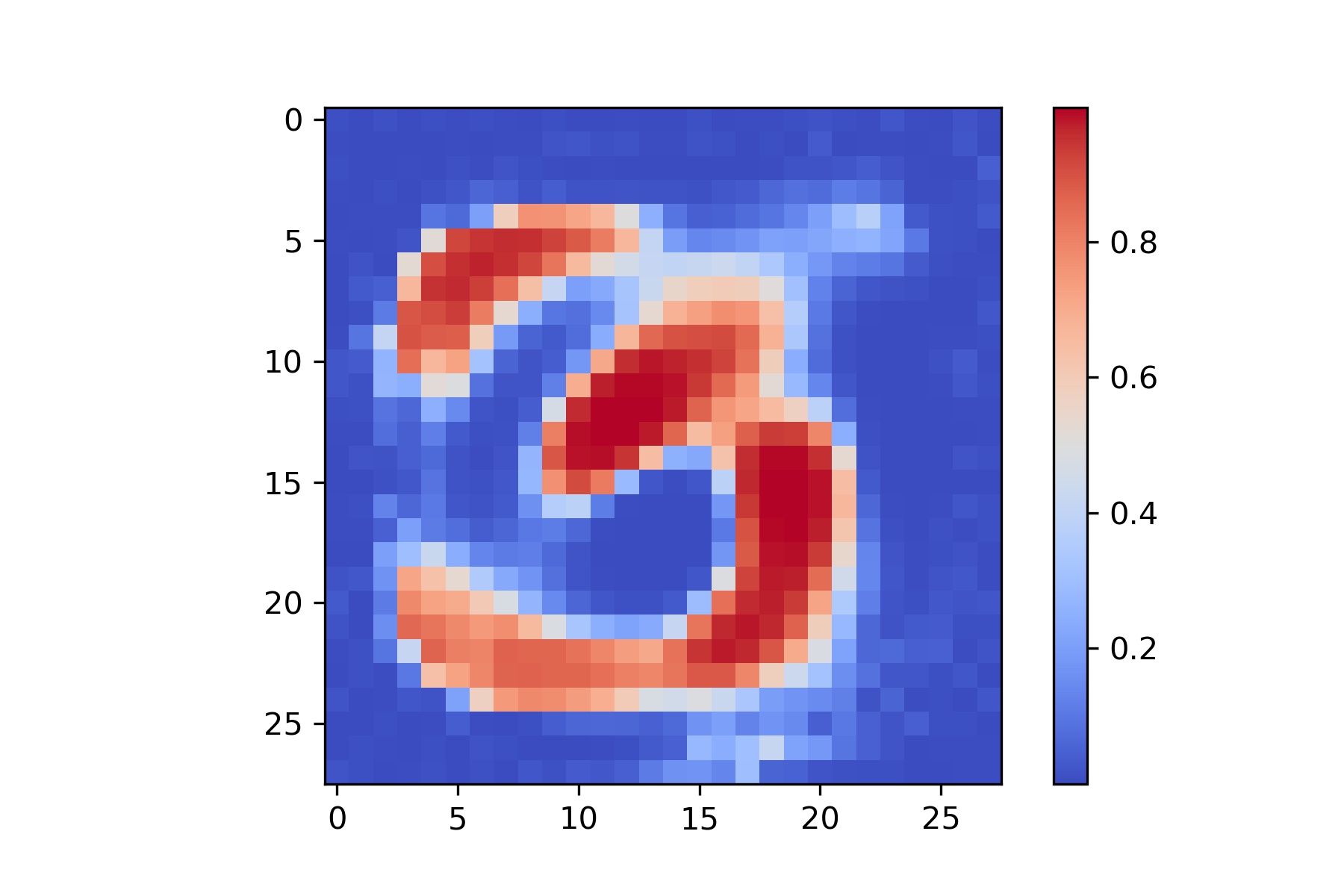}}
    \\
    \subfloat{\includegraphics[width=0.115\textwidth,trim={1.5cm 0 1.5cm 0},clip]{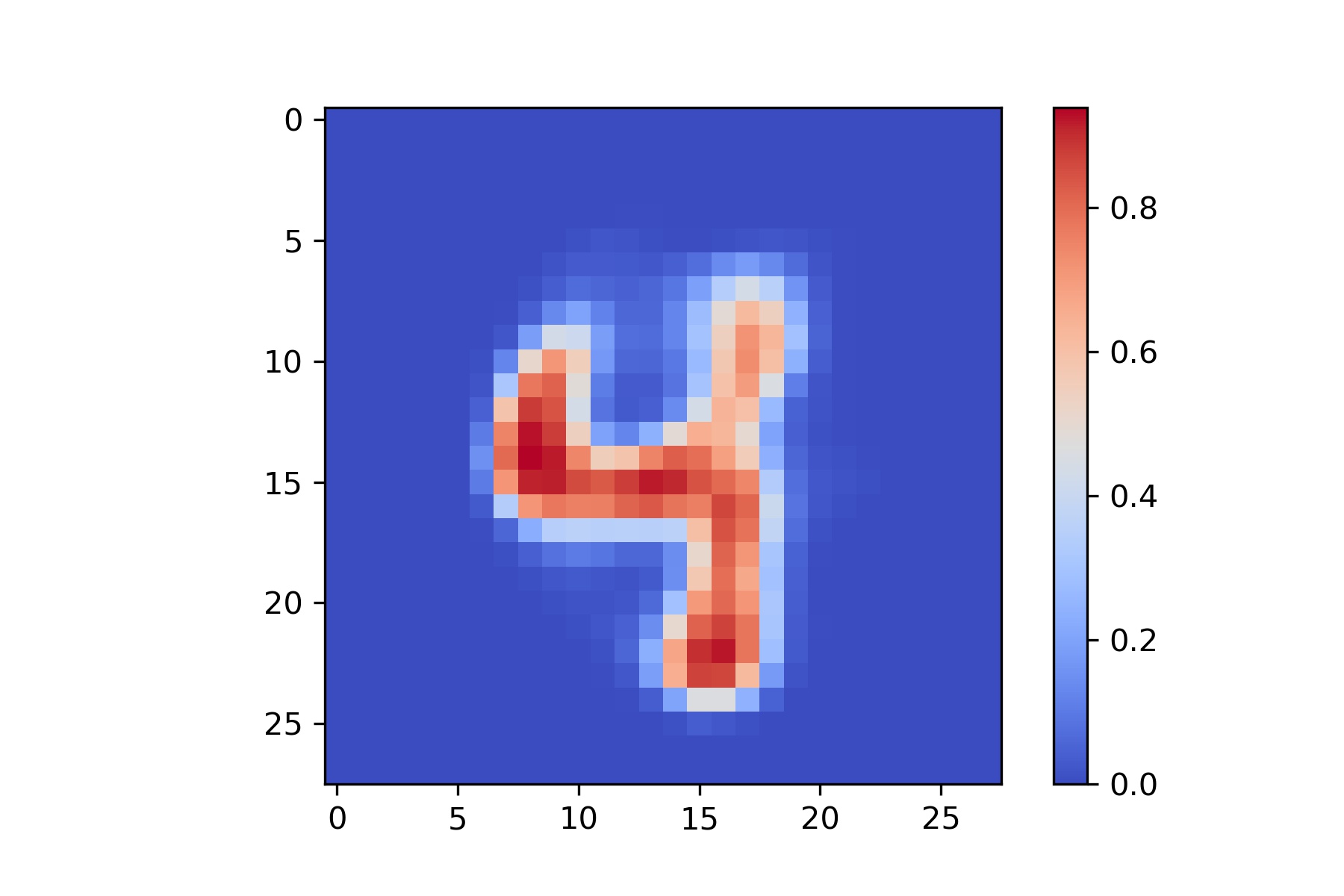}}
    \subfloat{\includegraphics[width=0.115\textwidth,trim={1.5cm 0 1.5cm 0},clip]{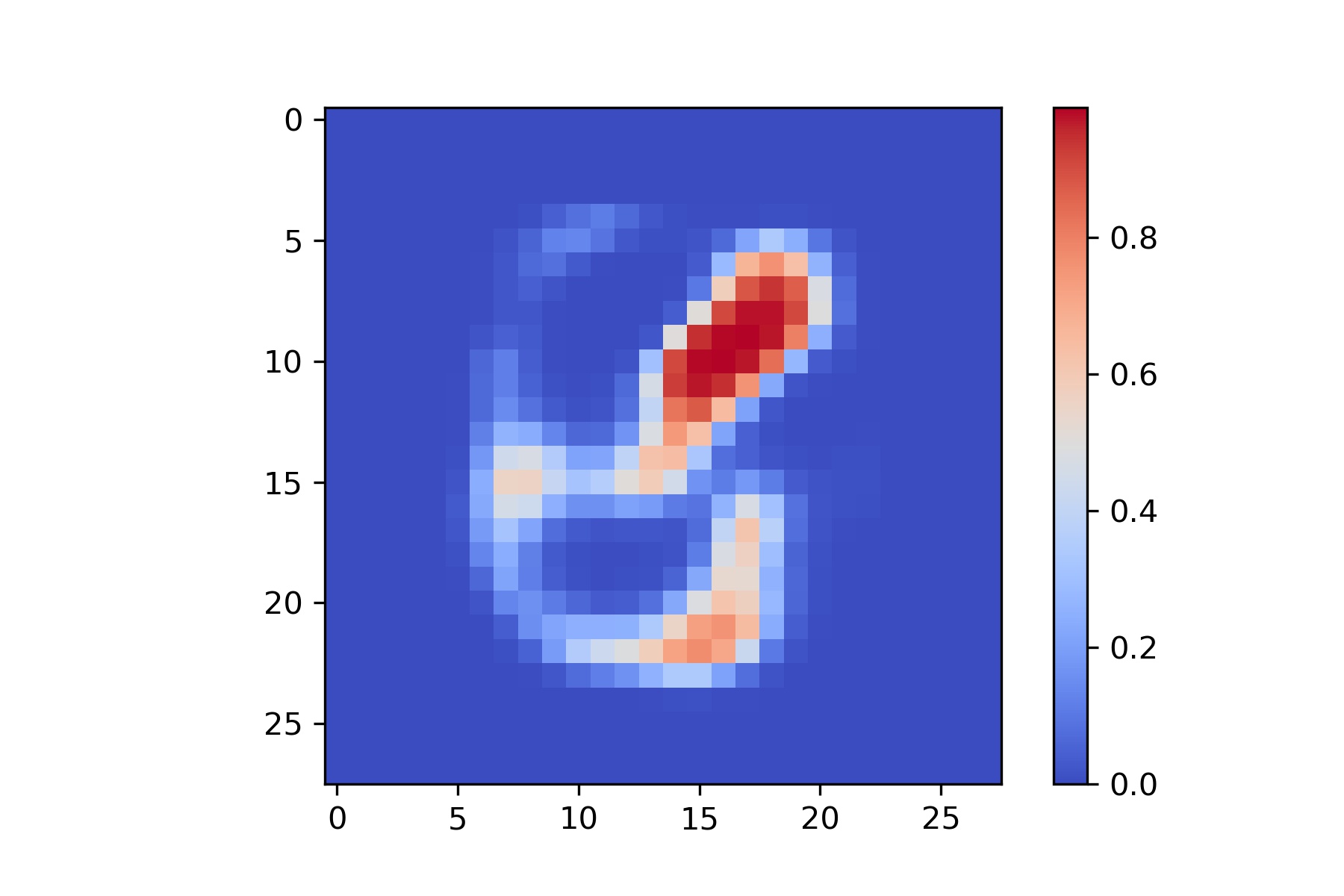}}
    \subfloat{\includegraphics[width=0.115\textwidth,trim={1.5cm 0 1.5cm 0},clip]{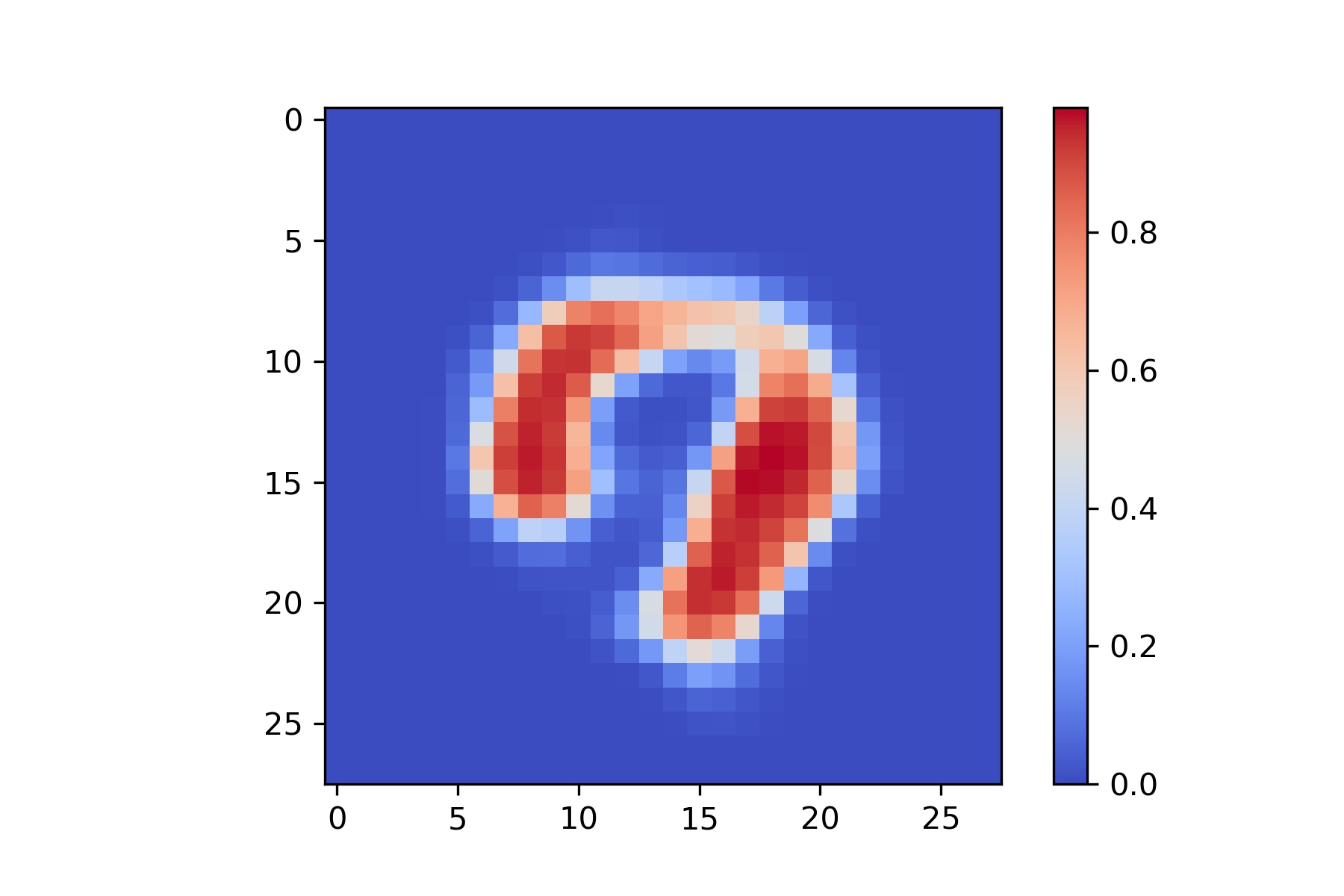}}
    \subfloat{\includegraphics[width=0.115\textwidth,trim={1.5cm 0 1.5cm 0},clip]{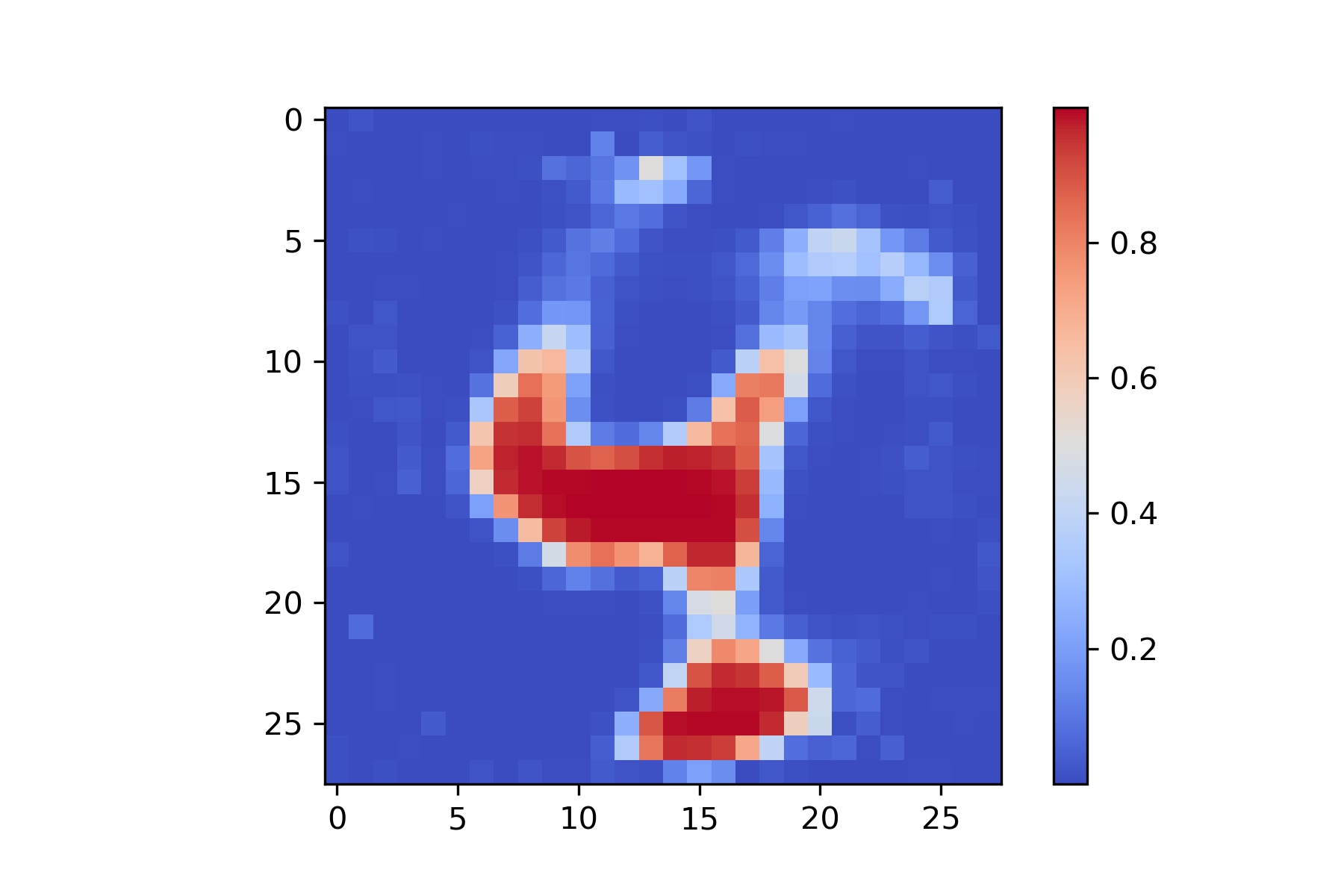}}
    \\
    
    \subfloat{\includegraphics[width=0.115\textwidth,trim={1.5cm 0 1.5cm 0},clip]{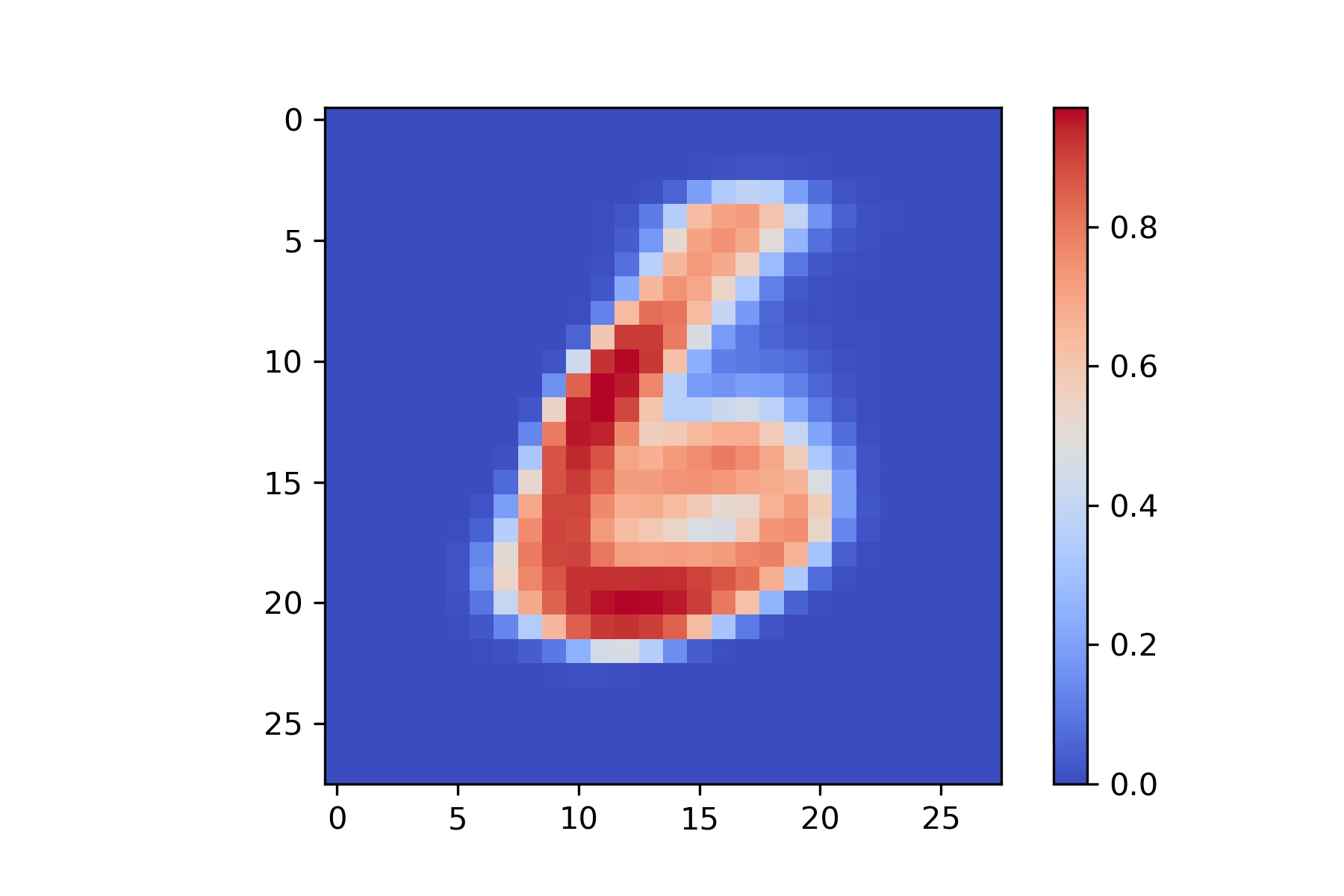}}
    \subfloat{\includegraphics[width=0.115\textwidth,trim={1.5cm 0 1.5cm 0},clip]{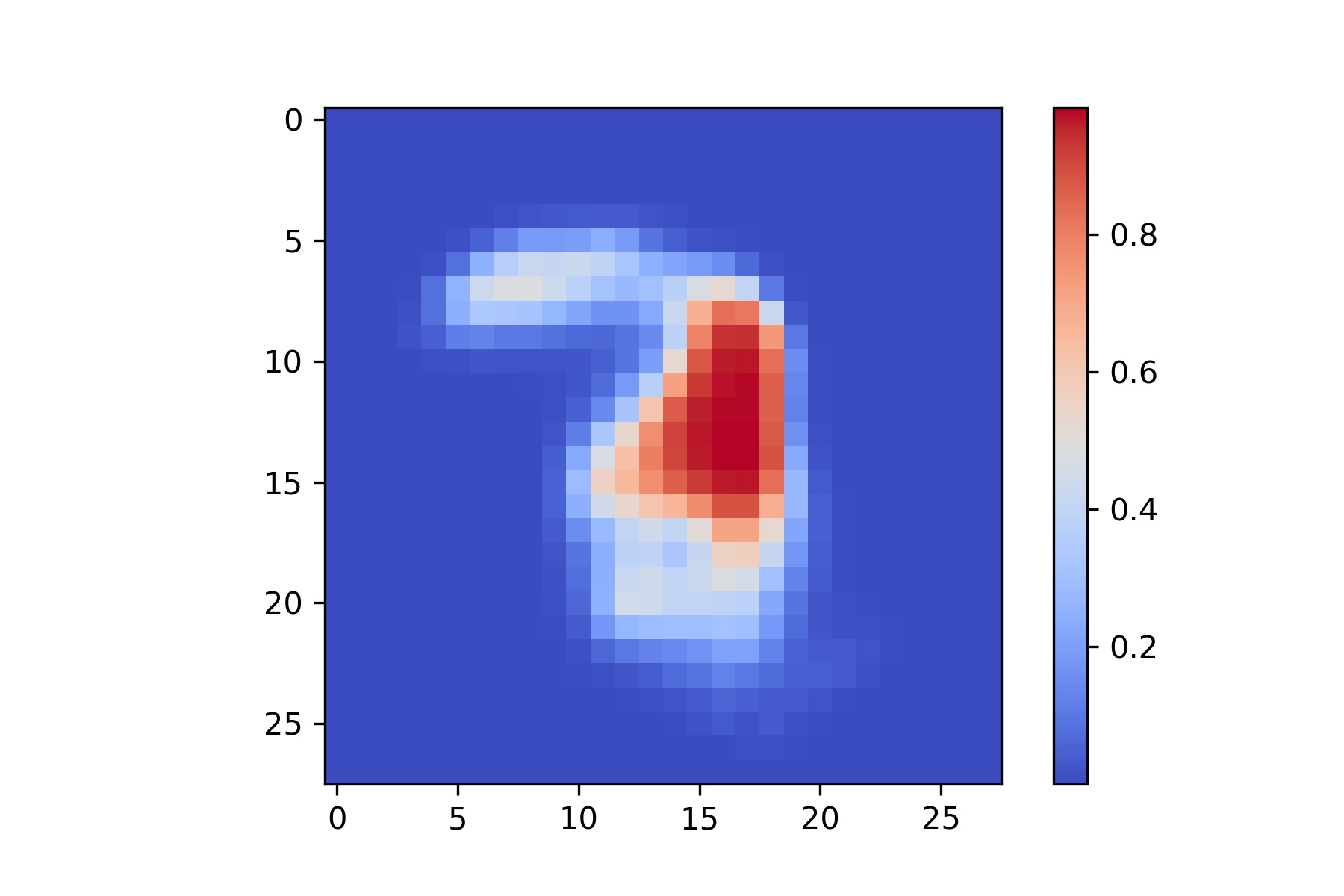}}
    \subfloat{\includegraphics[width=0.115\textwidth,trim={1.5cm 0 1.5cm 0},clip]{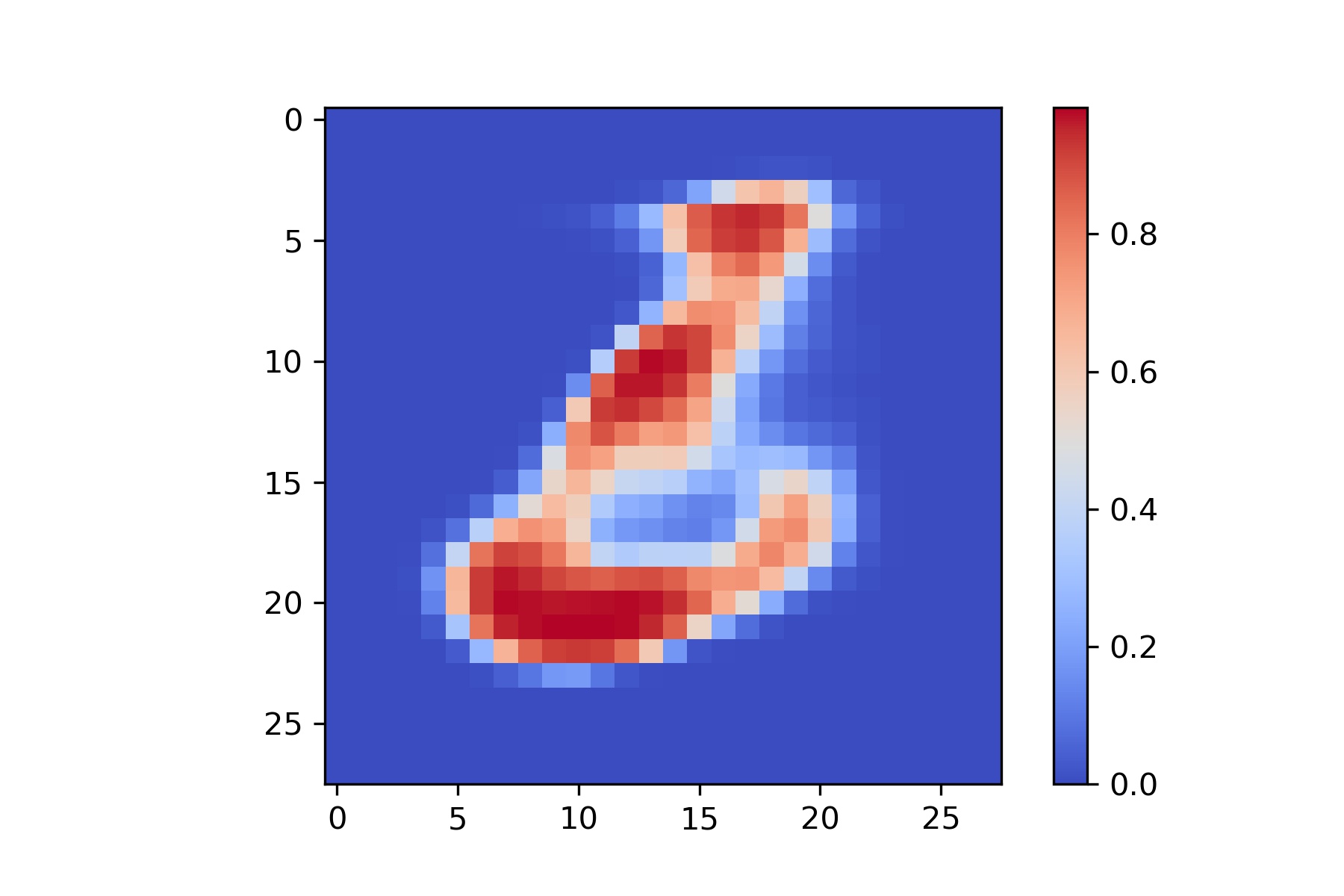}}
    \subfloat{\includegraphics[width=0.115\textwidth,trim={1.5cm 0 1.5cm 0},clip]{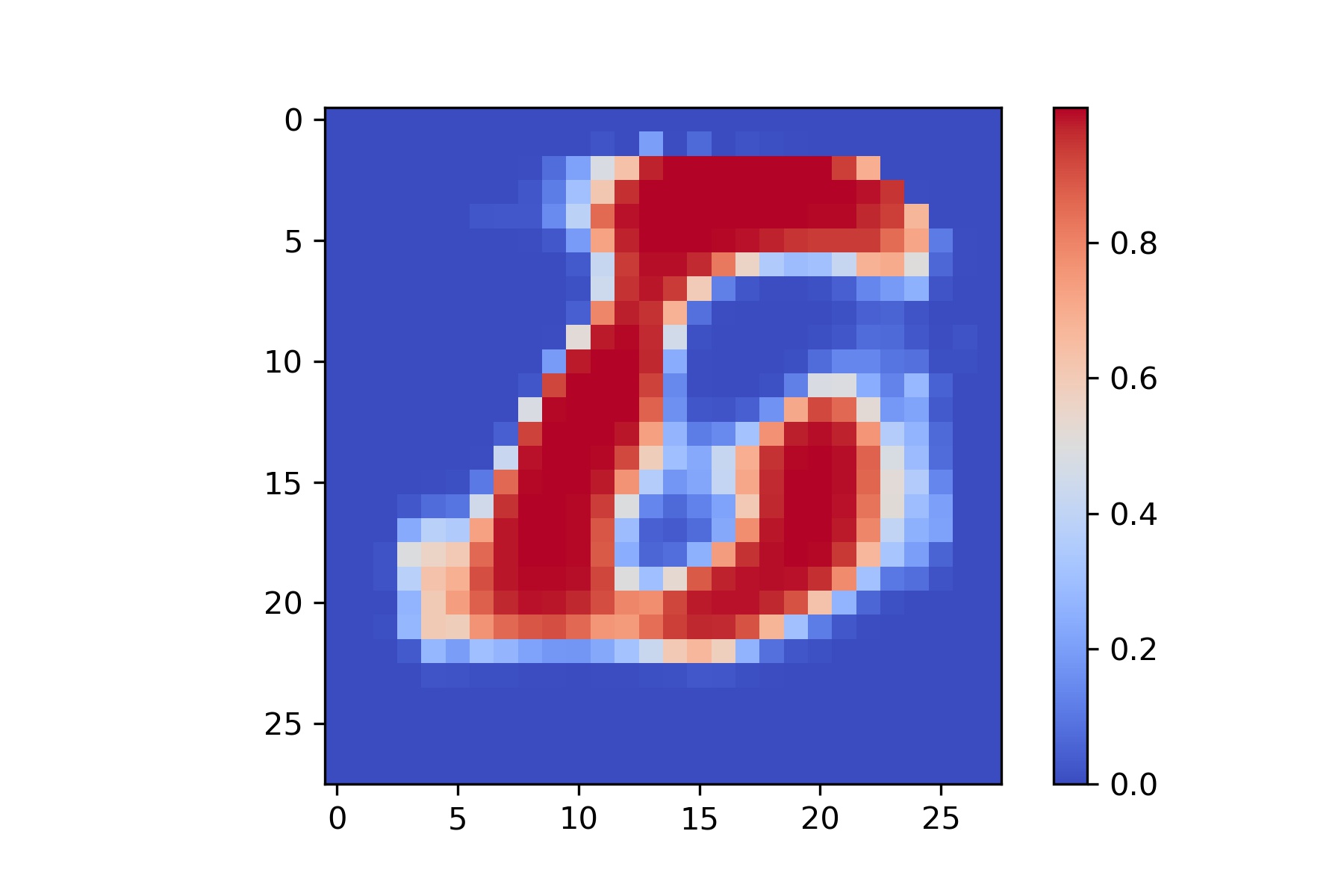}}
    \\
    \subfloat{\includegraphics[width=0.115\textwidth,trim={1.5cm 0 1.5cm 0},clip]{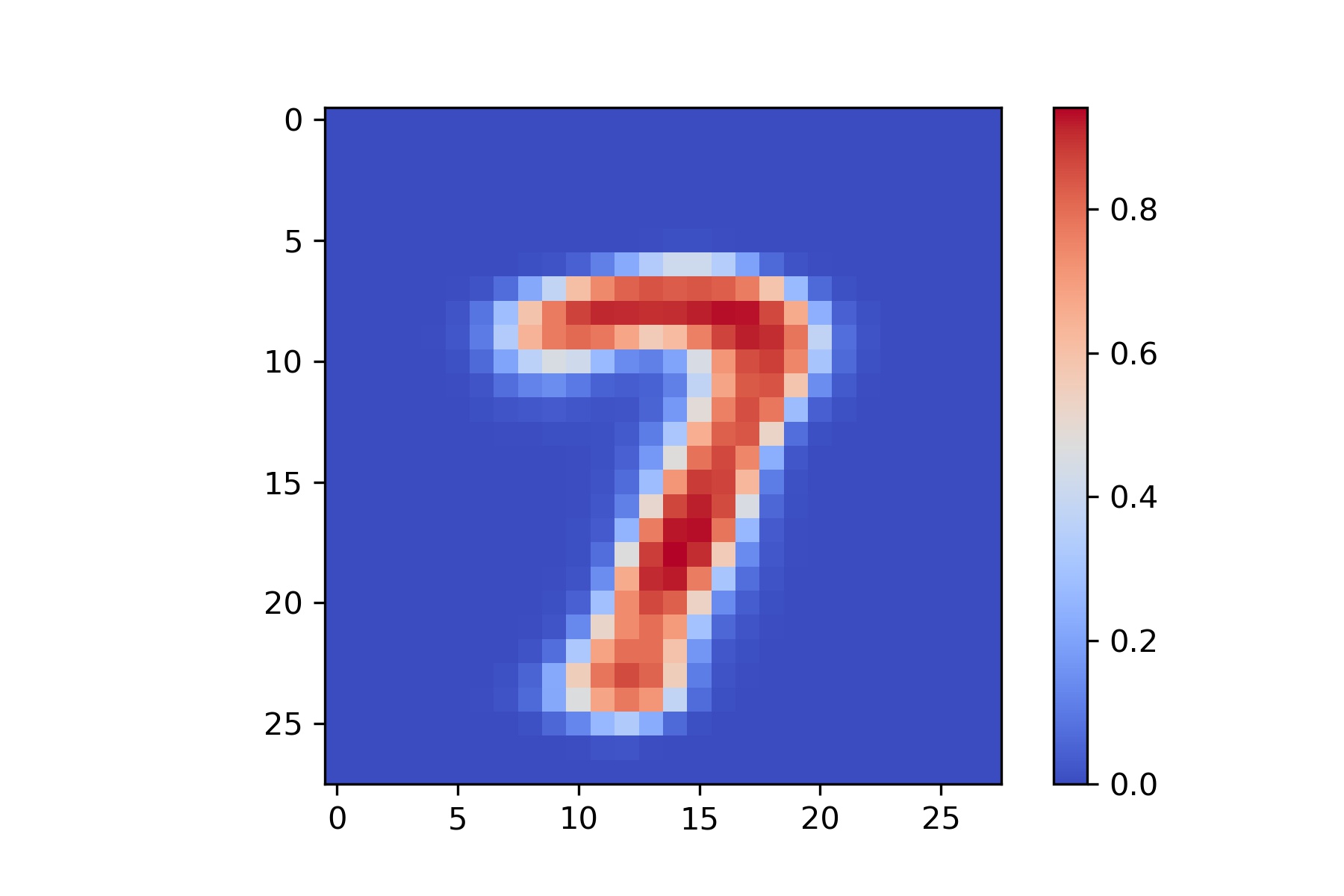}}
    \subfloat{\includegraphics[width=0.115\textwidth,trim={1.5cm 0 1.5cm 0},clip]{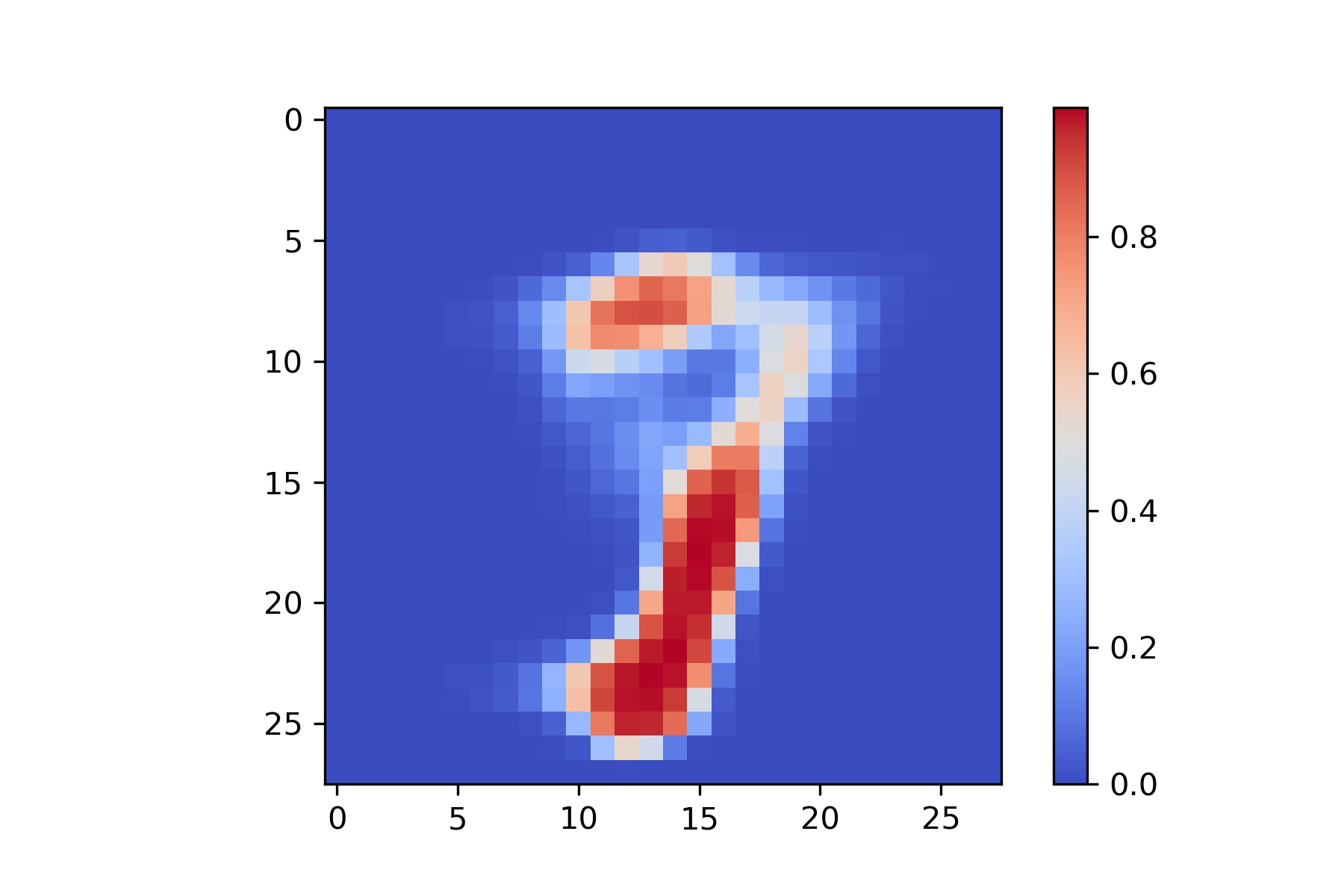}}
    \subfloat{\includegraphics[width=0.115\textwidth,trim={1.5cm 0 1.5cm 0},clip]{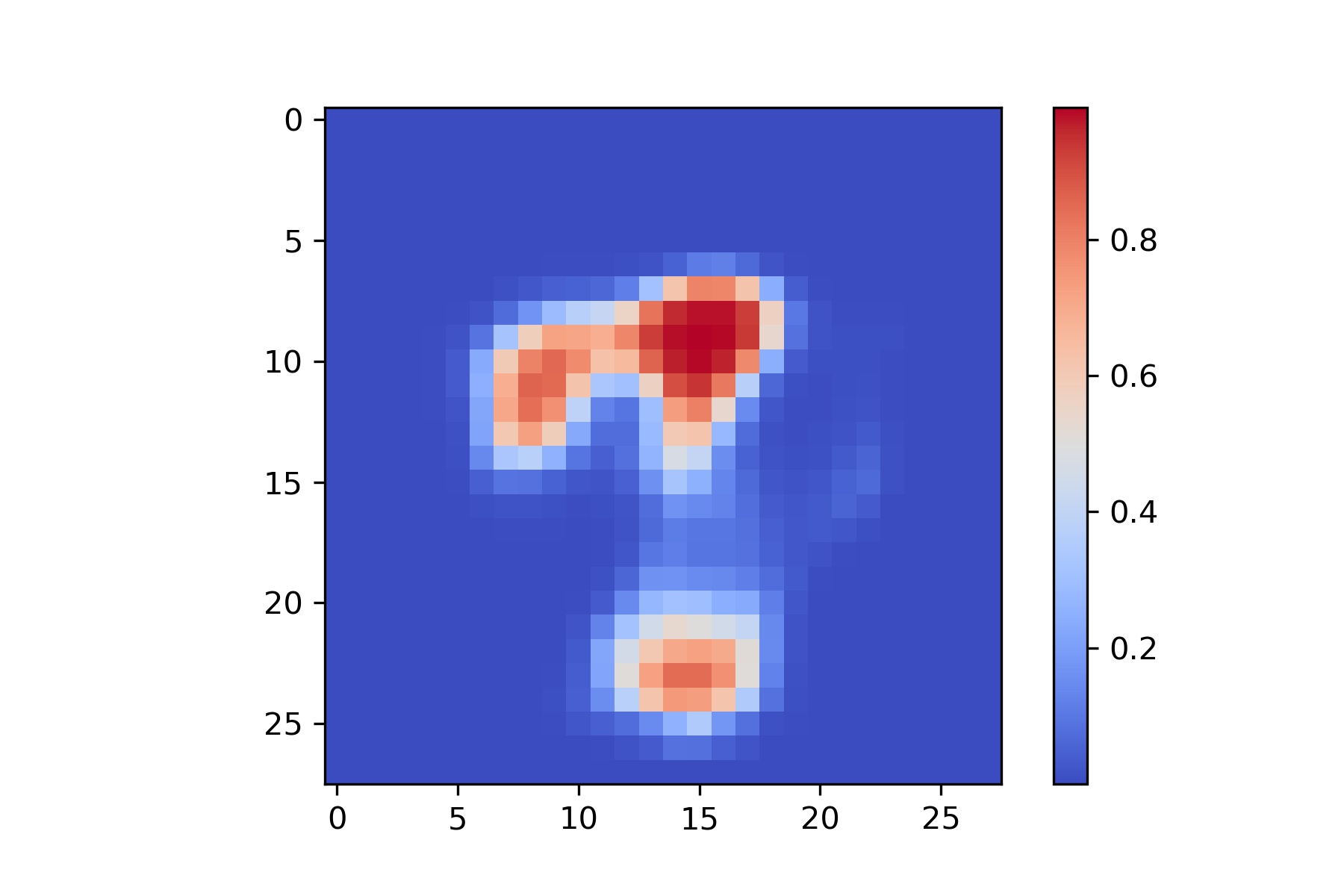}}
    \subfloat{\includegraphics[width=0.115\textwidth,trim={1.5cm 0 1.5cm 0},clip]{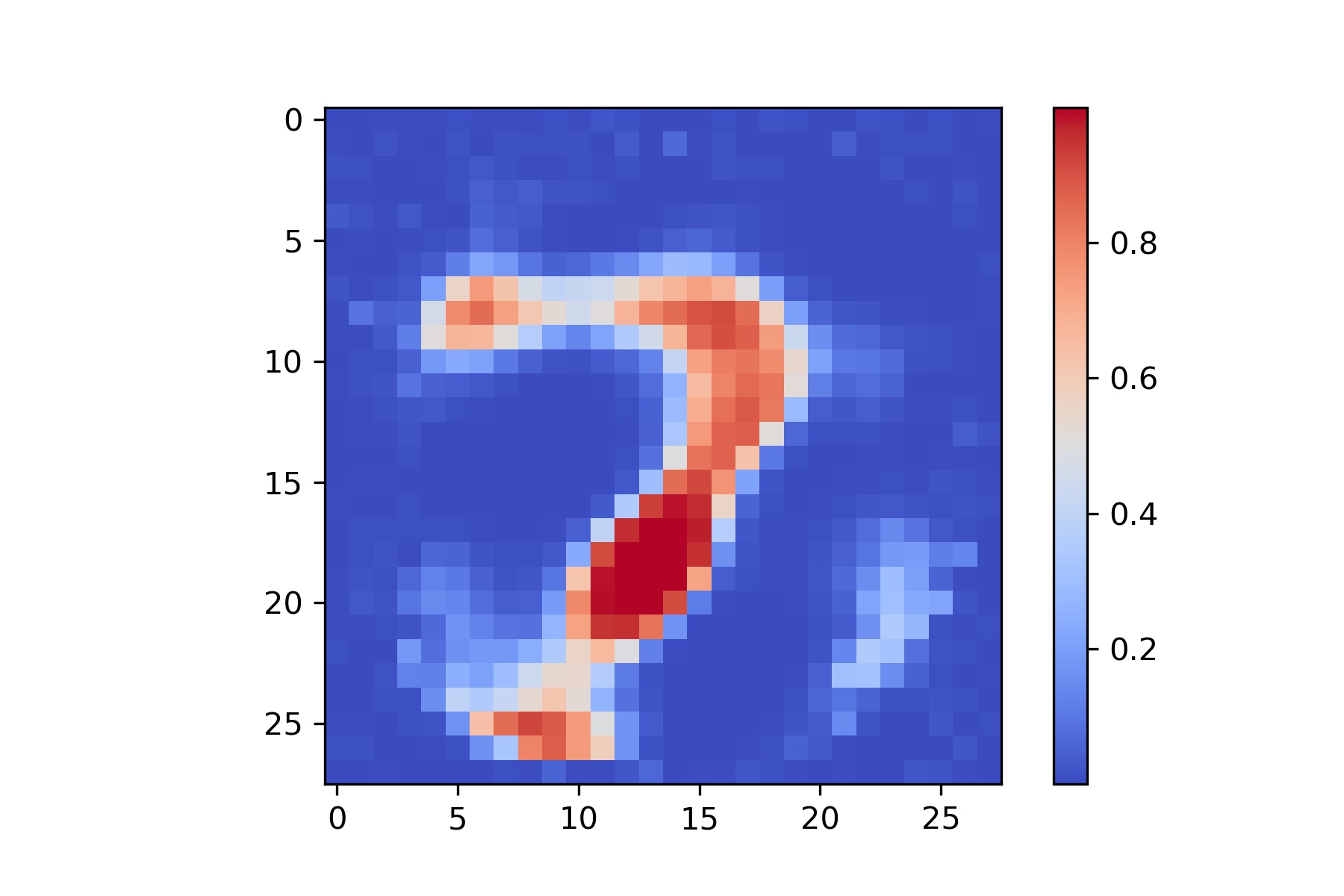}}
    \\
    \setcounter{subfigure}{0}
    \subfloat[]{\includegraphics[width=0.115\textwidth,trim={1.5cm 0 1.5cm 0},clip]{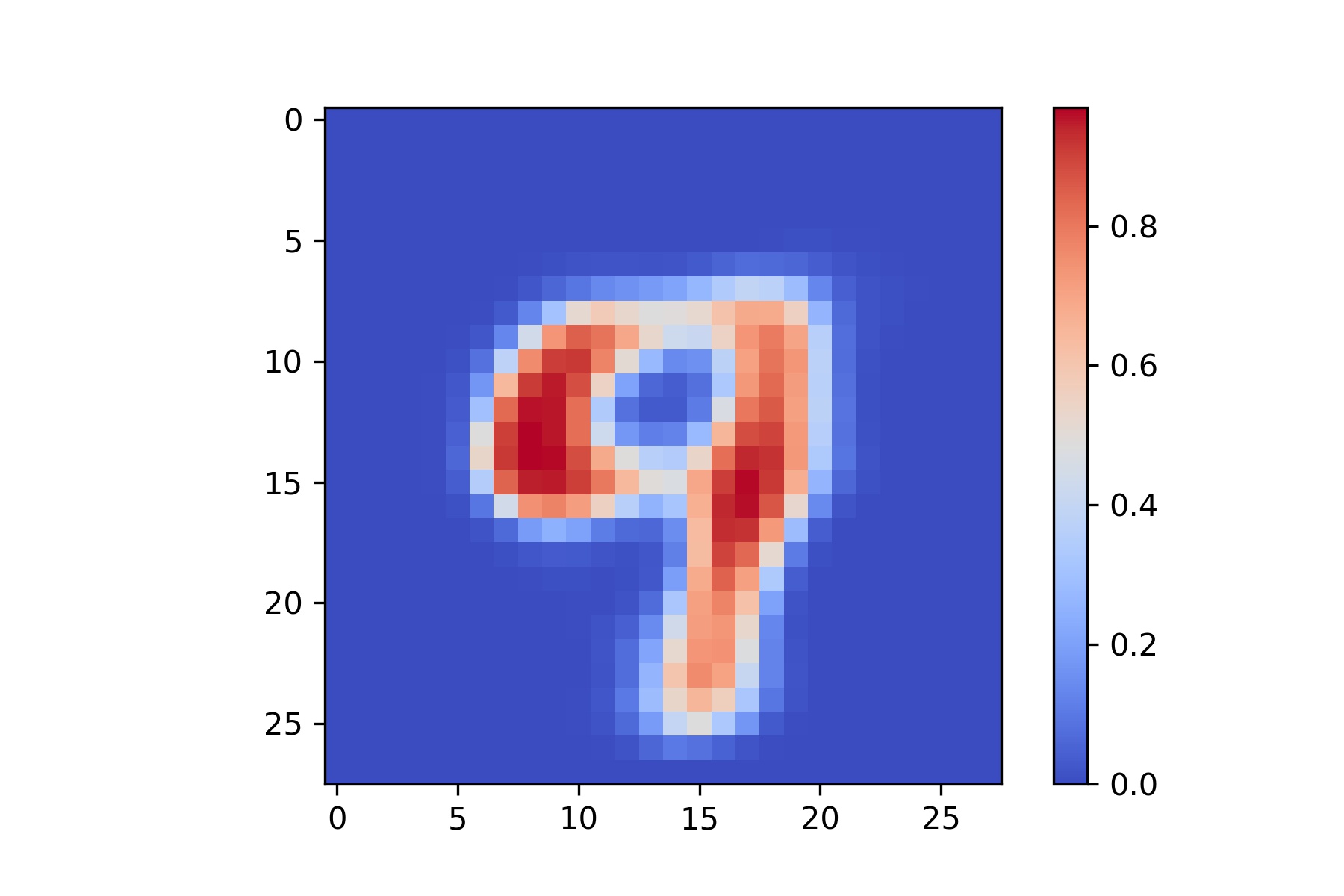}}
    \subfloat[]{\includegraphics[width=0.115\textwidth,trim={1.5cm 0 1.5cm 0},clip]{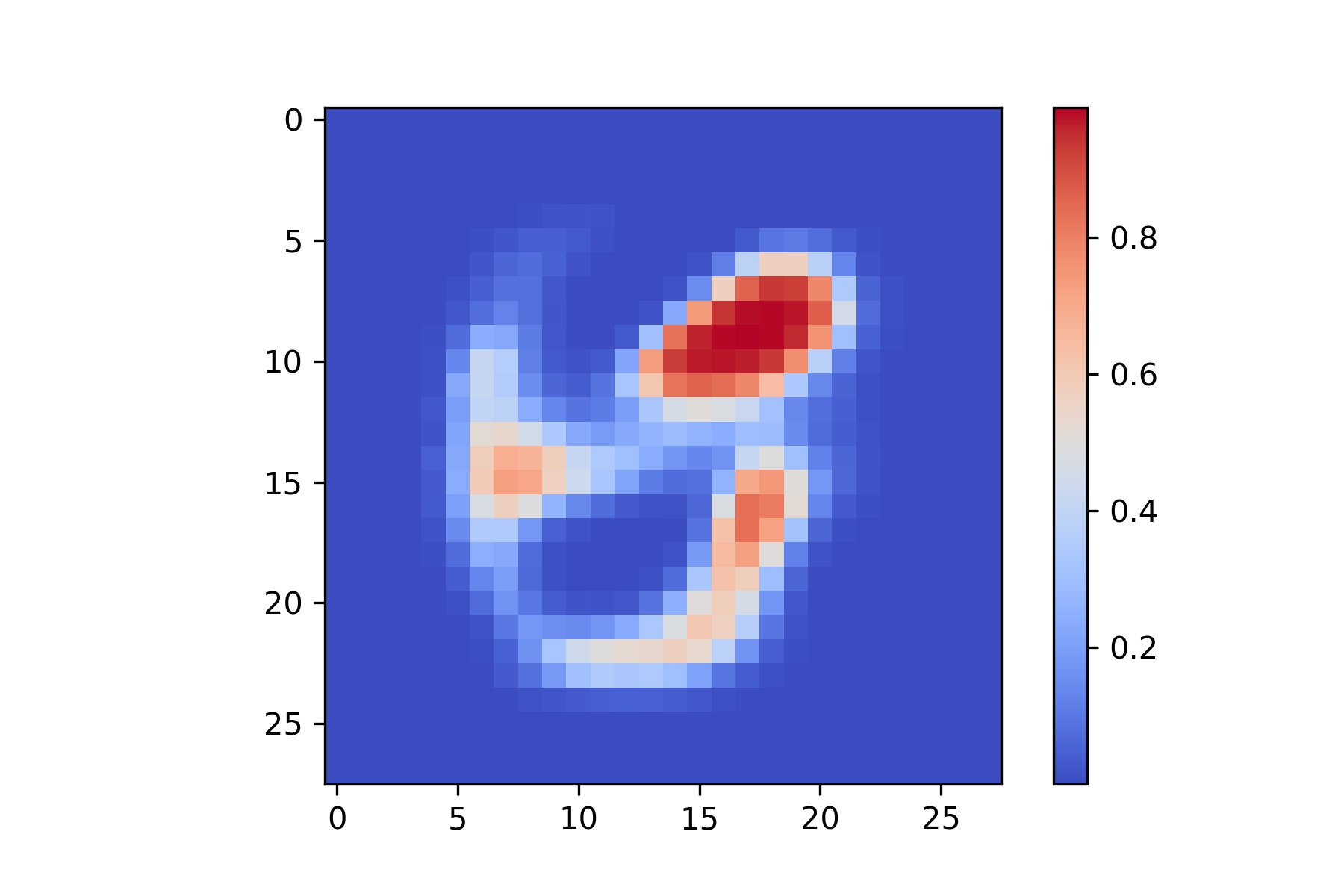}}
    \subfloat[]{\includegraphics[width=0.115\textwidth,trim={1.5cm 0 1.5cm 0},clip]{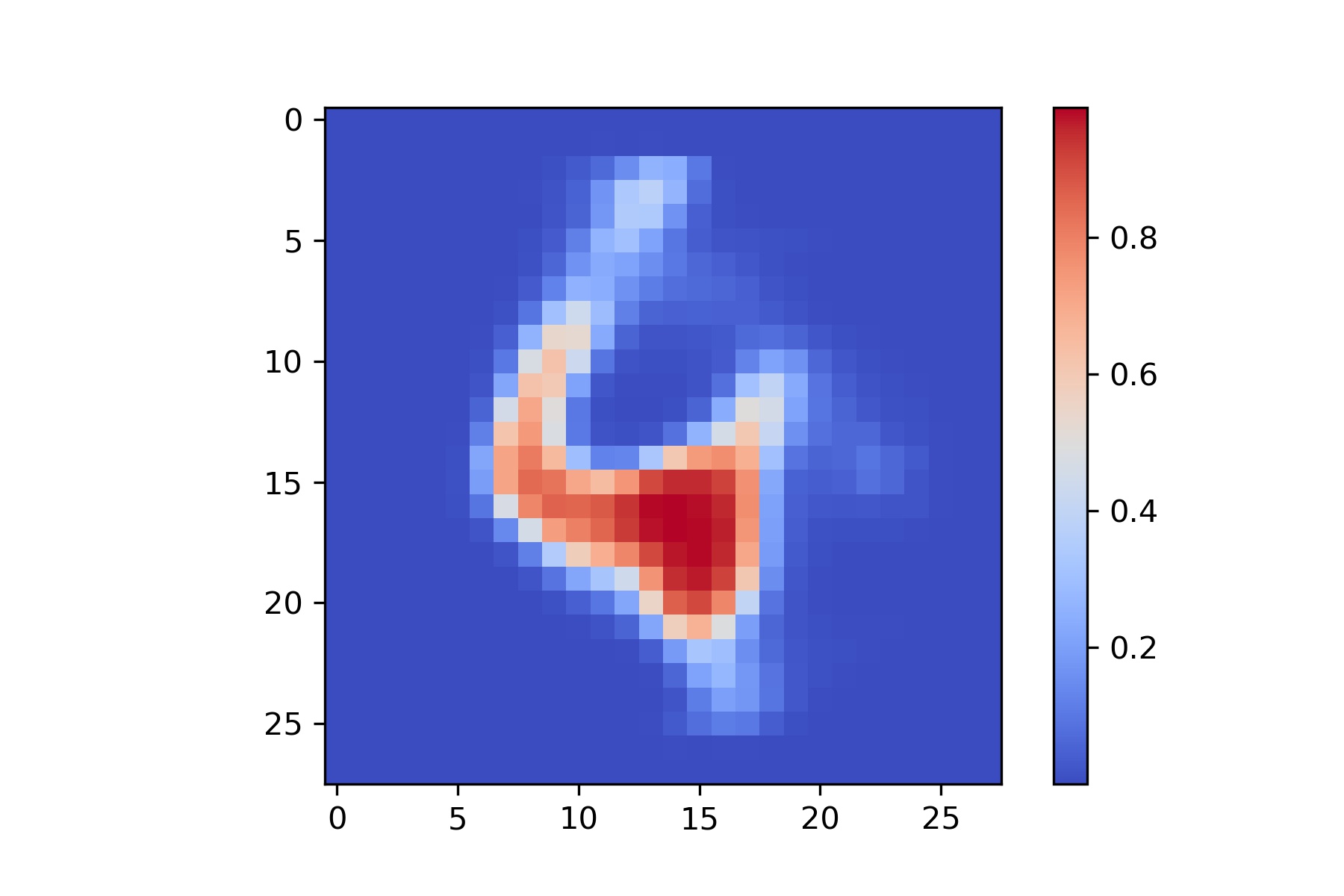}}
    \subfloat[]{\includegraphics[width=0.115\textwidth,trim={1.5cm 0 1.5cm 0},clip]{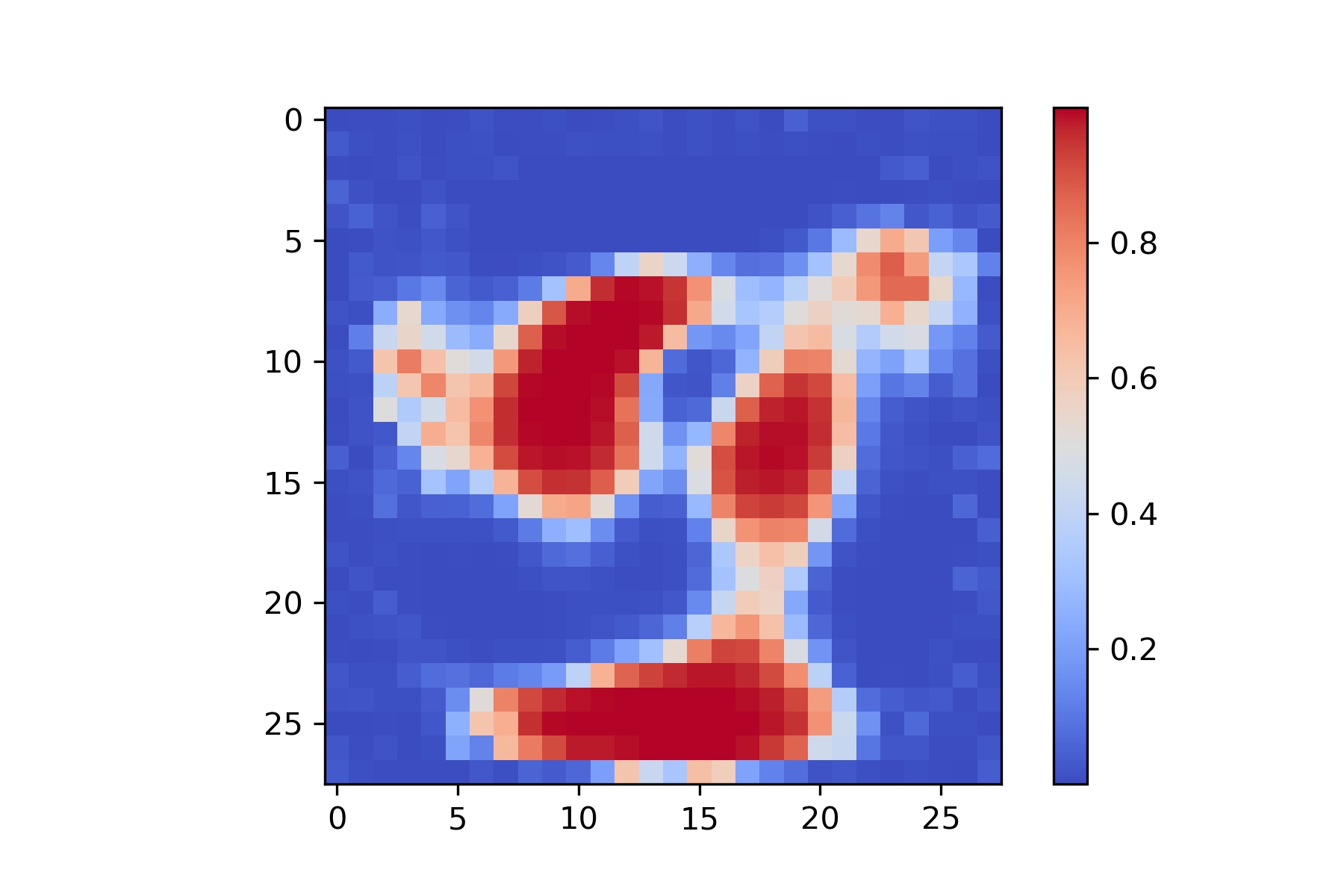}}
    \\
    
    \caption{Results of our experiments on the MNIST data set. (a) Images reconstructed using the Auto-Encoder. (b) and (c) Images reconstructed after intervening on each causal latent variable. (d) Images reconstructed using only the causal subset of latent variables, having set all other latent variables to zero.}
    \label{fig:mnist}
\end{figure}

We show the results of our experiments in Figure \ref{fig:mnist} (best viewed in color). We observe that intervening on and zeroing out each of the identified latent variables with large causal attribution for the predicted CNN label, indeed results in a significant distortion of the reconstructed image relative to the original image. The distorted images can be viewed in columns (b) and (c), while the original image can be viewed in column (a) of Figure \ref{fig:mnist}. On the other hand, images reconstructed using only the identified causal latent variables end up being similar to the original image. The images reconstructed using only the identified causal latent variables are shown in column (d) of Figure \ref{fig:mnist}. Interestingly, we observed that all of the causal effect estimation methods consistently agreed with each other in identifying the causal latent variables in this experiment. These results demonstrate that our method is indeed effective in correctly interpreting black box predictive models via causal attribution. 

\noindent{\bf Example:} Suppose a user wants to know why a specific image is classified as the digit 3. The answer to this question follows from the fourth row of Figure \ref{fig:mnist}, which shows that the variables $Z_2$ and $Z_7$ (which we identified to be the responsible causal latent variables for class label 3) have large causal attributions for that prediction outcome. If we set the value of any of these variables to zero (shown in columns (b) and (c) of the figure respectively for $Z_2$ and $Z_7$ for the case of digit 3), the image will no longer look like the digit 3. If we set the values of all of the latent variables except  $Z_2$ and $Z_7$ to zero, we note that even though the image is distorted, we can recognize digit 3. These results show that the specific image was classified as digit 3 because of the large causal effect of the latent variables $Z_2$ and $Z_7$.

\subsubsection{{\bf Parkinson's Disease Telemonitoring data set}}
Table \ref{table:parkinson} shows our estimates of the causal effects of all 18 features of the data set on the UPDRS predicted by the neural network.\footnote{We note that gender is the only discrete attribute in this data set and we have treated it as binary in our experiments. This means that the change in output is proportionate to $\exp{(\hat{\mu})}$ in the case of gender, where $\hat{\mu}$ is the estimated causal effect of gender on UPDRS.} Interestingly, we observe that  age and gender are found to have large causal effects on the predicted UPDRS score. Specifically, predicted UPDRS increases with age, and is lower in females as compared to males. Interestingly, we find that biomedical voice measures, Jitter (Abs), HNR, DFA, and PPE have significant causal effects on predicted UPDRS score (with mostly a general agreement among the different causal effect estimation methods). We note that these findings are consistent with the relevant literature \cite{diederich2003parkinson,tsanas2009accurate}, thus supporting the notion that causal attribution of predictions of black box models can provide useful insights into data.

\begin{table*}[t] 
\caption{\label{table:parkinson} Estimates and p-value significance of the causal effect of each feature on the predicted UPDRS by the deep neural network on the PD Telemonitoring data set.}
\resizebox{\textwidth}{!}{\begin{tabular}{lcccccccccccc}
\toprule
& \multicolumn{2}{c}{CBPS} & \multicolumn{2}{c}{NPCBPS} & \multicolumn{2}{c}{PSWGBM} & \multicolumn{2}{c}{OPTWEIGHT} & \multicolumn{2}{c}{IPTW} & \multicolumn{2}{c}{SUPER} \\ \midrule
Feature        & Est.    & P    & Est.     & P     & Est.    & P   & Est.       & P      & Est.    & P   & Est.     & P    \\
\midrule
Age             & 16.12       & \textless{}0.01       & 15.20        & \textless{}0.01        & 15.65       & \textless{}0.01      & 16.81          & \textless{}0.01         & 15.56       & \textless{}0.01       & 15.54        & \textless{}0.01       \\
Gender             & -1.48       & \textless{}0.01       & -1.76        & \textless{}0.01        & -1.51       & \textless{}0.01      & -2.37          & \textless{}0.01         & -1.83       & \textless{}0.01       & -1.83        & \textless{}0.01       \\
Jitter (\%) & 62.75       & 0.12       & 16.99        & 0.70        & 12.20       & 0.77      & -11.30         & 0.78         & -5.07       & 0.68       & -6.80        & 0.58       \\
Jitter (Abs)     & -13.21      & \textless{}0.01       & -38.50       & \textless{}0.01        & -16.84      & 0.02      & -60.94         & \textless{}0.01         & -36.24      & \textless{}0.01       & -36.16       & \textless{}0.01       \\
Jitter:RAP     & -2713.96    & \textless{}0.01       & 23826.27     & \textless{}0.01        & -3561.39    & 0.21      & 3054.57        & 0.38         & -4775.70    & \textless{}0.01       & -4803.89     & \textless{}0.01       \\
Jitter:PPQ5    & 29.17       & 0.02       & 21.58        & 0.39        & -31.16      & 0.19      & 57.19          & 0.04         & -43.32      & \textless{}0.01       & -43.54       & \textless{}0.01       \\
Jitter:DDP     & 2536.11     & 0.01       & 2307.36      & 0.49        & 3591.26     & 0.20      & -3029.24       & 0.38         & 4850.61     & \textless{}0.01       & 4879.72      & \textless{}0.01       \\
Shimmer         & 117.01      & \textless{}0.01       & -13.76       & 0.74        & 49.15       & 0.10      & -102.39        & 0.03         & 62.88       & \textless{}0.01       & 59.94        & \textless{}0.01       \\
Shimmer:dB     & 0.61        & 0.95       & -16.65       & 0.40        & -17.11      & 0.35      & -8.79          & 0.65         & 36.93       & \textless{}0.01       & 37.72        & \textless{}0.01       \\
Shimmer:APQ3   & 5119.44     & \textless{}0.01       & -1094.98     & 0.53        & 1628.27     & 0.45      & 455.96         & 0.72         & -940.25     & 0.07       & -938.57      & 0.07       \\
Shimmer:APQ5   & -34.58      & \textless{}0.01       & -43.04       & 0.07        & -16.91      & 0.30      & -75.33         & \textless{}0.01         & -44.94      & \textless{}0.01       & -45.67       & \textless{}0.01       \\
Shimmer:APQ11  & 32.02       & 0.05       & 48.73        & 0.03        & 12.18       & 0.38      & 62.88          & \textless{}0.01         & 22.24       & \textless{}0.01       & 22.67        & \textless{}0.01       \\
Shimmer:DDA    & -3723.52    & \textless{}0.01       & 210.39       & 0.91        & -1674.89    & 0.44      & -348.83        & 0.79         & 1065.98     & 0.04       & 1064.20      & 0.04       \\
NHR             & 0.07        & 0.99       & 26.65        & 0.01        & -5.42       & 0.55      & -36.64         & \textless{}0.01         & -8.15       & 0.09       & -8.00        & 0.10       \\
HNR             & -23.40      & \textless{}0.01       & -29.40       & \textless{}0.01        & -18.58      & \textless{}0.01      & -28.17         & \textless{}0.01         & -35.13      & \textless{}0.01       & -35.43       & \textless{}0.01       \\
RPDE            & 3.30        & 0.19       & 3.77         & 0.14        & -1.11       & 0.65      & 3.13           & 0.24         & -0.99       & 0.69       & -1.05        & 0.67       \\
DFA             & -7.14       & \textless{}0.01       & -6.59        & \textless{}0.01        & -9.85       & \textless{}0.01      & -2.97          & 0.04         & -7.09       & \textless{}0.01       & -7.10        & \textless{}0.01       \\
PPE             & 12.72       & \textless{}0.01       & 18.00        & \textless{}0.01        & 21.17       & \textless{}0.01      & 17.02          & \textless{}0.01         & 10.15       & \textless{}0.01       & 9.96         & \textless{}0.01       \\ \bottomrule
\end{tabular}
}
\end{table*}

\section{Summary and Discussion} \label{discussion}
\subsection{Summary}
Predictive models trained using machine learning are extensively used across a wide range of high-stakes applications including health care, security, criminal justice, finance, and education. Such utilization imposes a pressing need for effective techniques that can explain the predictive models and their predictions. We have addressed this problem in settings where the predictive model is a {\em black box}, meaning that users can only observe the response of the model to various inputs, but have no information about the structure of the predictive model, its parameters, or the objective function and the algorithms that are used to optimize the model. We reduce the problem of interpretation of black box predictive models to the problem of estimating the causal effects of model inputs on model outputs from  observations of outputs  of the model for a sufficiently large and diverse sample of inputs. To the best of our knowledge, we offer the first model agnostic solution to interpretation of black box predictive models via causal attribution.  In contrast to the only existing approach to interpretation via causal attribution \cite{chattopadhyay2019neural}, our solution does not require the interpretation algorithm to have access to the internal structure, parameters, or the objective function and the algorithm used to optimize the  black box predictive model. Hence, the approach can be applied, in principle, to {\em any} black box predictive model, so long as it can probe the model and  observe the model's output for any supplied input data sample. We estimate the causal effects of model inputs on model output using state-of-the art variants of the Rubin Neyman {\em potential outcomes} framework for estimating causal effects from observational data. The proposed solution works for both discrete as well as continuous valued inputs. We show how the resulting causal attribution of responsibility for model predictions to a subset of the inputs, can be used to explain the observed differences in the model's outputs in different cases, e.g.,``Why did the model recommend that John's loan application be approved but Sarah's was not?'' We demonstrate the effectiveness of our approach to the interpretation of black box predictive models  via causal attribution  using deep neural network models trained on one synthetic data set (where the input variables that impact the  output variable are known by design) and two real-world data sets: Handwritten digit  classification, and  Parkinson's disease severity prediction. 

\subsection{Related Work}
The nature and desiderata of explanations have been widely studied in philosophy of science, cognitive science, and social sciences \cite{kitcher1987van,salmon1984scientific,salmon1998causality,kass1988need,miller2019explanation}. The ability to interpret a predictive model is a necessary condition for being able to explain it.  A dominant approach to model interpretation involves {\em attribution} of responsibility for the model output to the model's inputs (i.e., features) \cite{sundararajan2017axiomatic,roscher2020explainable}.

Shapley values \cite{shapley1953value}  offer an alternative approach to explaining  predictive models \cite{datta2016algorithmic,lipovetsky2001analysis,vstrumbelj2014explaining,lundberg2017unified,ancona2019explaining}. In this setting, a  prediction is explained by assuming that each feature  of the data sample is a ``player'' in a game where the payout is the difference between the predicted output for that feature, and the average predicted output over the entire data set. Techniques from game theory are used to  fairly distribute the ``payout'' among the features \cite{shapley1953value}. The Shapley value of a feature is the average marginal contribution of a feature value across all possible coalitions. While Shapley values provide an intuitive approach to scoring features, because of the combinatorial nature of the task, the computational cost of calculating Shapley values becomes prohibitive when the number of features is large \cite{ancona2019explaining}. Recent work has pointed out that mathematical problems arise when Shapley values are used for feature importance and that the solutions to mitigate these necessarily induce further complexity \cite{kumar2020problems}. In light of this, causal attribution methods introduced in this paper offer an attractive alternative to Shapley values for assessing feature importance. 

As previously mentioned, much work has focused on attribution methods for interpreting predictive models (reviewed in \cite{mueller2019explanation,guidotti2018survey,adadi2018peeking,montavon2018methods}), also known as {\em post hoc} interpretations \cite{guidotti2018survey,lipton2016mythos}. Such methods include techniques for visualizing the effect of the model inputs on its outputs \cite{simonyan2013deep,yosinski2015understanding,zeiler2014visualizing,letham2015interpretable}, methods for extracting purportedly human interpretable rules from black box models \cite{andrews1995survey,frosst2017distilling,setiono1995understanding,towell1993extracting,thrun1995extracting}, feature scoring methods  that assess the importance of individual features on the prediction \cite{friedman2001greedy,goldstein2015peeking,datta2016algorithmic,lipovetsky2001analysis,vstrumbelj2014explaining,lundberg2017unified,ancona2019explaining,chen2018learning}, gradient based methods  that assess how changes in inputs impact the model predictions \cite{baehrens2010explain,shrikumar2016not,shrikumar2017learning,bach2015pixel,chen2018learning}, and techniques for approximating local decision surfaces in the neighborhood of the input sample via localized regression \cite{ribeiro2016should,selvaraju2017grad,bhatt2020explainable}. A shared feature of all of these model interpretation methods is that they focus primarily on how a model's inputs correlate with its outputs. They often fail to generate reliable attributions \cite{sundararajan2017axiomatic,kindermans2019reliability}, let alone interpretations that support explanations \cite{mueller2019explanation,guidotti2018survey,adadi2018peeking,montavon2018methods}. 

On the other hand, the causal underpinnings of explanations have been well-recognized in the philosophy of science, cognitive science, and social sciences literature \cite{salmon1984scientific,salmon1998causality,kitcher1987van,kass1988need,miller2019explanation,van1988pragmatic}. There is a growing realization that explanations in general, and those of predictive models and their predictions in particular, have to be necessarily causal  \cite{halpern2005causes,mueller2019explanation,miller2019explanation}. Purely correlation based methods fail to adjust for the effect of confounders and hence are incapable of providing causal attributions. In contrast, causal inference methods  \cite{pearl2009causality,hernan2020causal} offer powerful machinery for causal attribution.

Our work is inspired by the seminal work of \cite{chattopadhyay2019neural} who offered the first causal attribution method for deep neural networks by estimating the causal effect of each of the  network inputs on the network output. They achieve this by first translating the deep neural network into a functionally equivalent Causal Bayesian Network  \cite{pearl2009causality} which is then used to calculate the relevant causal effects. A key limitation of their method is that it requires the causal attribution algorithm to have access to the structure as well as parameters of the trained deep neural network. Our approach overcomes this limitation by observing that the Causal Bayesian Network obtained from a deep neural network, or for that matter, any predictive model trained on a given data set, fundamentally cannot provide any information that is not available in the data used to train the network. Hence, assuming that the trained network is sufficiently accurate on the training data, it should be possible to estimate the causal effects of the model inputs on its outputs using state-of-the-art methods for estimating causal effects from observational data. 

We would be amiss if we did not mention possible connections between our approach and the well-known Partial Dependence Plots (PDP) \cite{friedman2001greedy}. PDPs offer a technique for visualizing the marginal effect of features on the output of a predictive model. Suppose that $g(x)$ is a predictive model trained using features in ${\bf X}$. Suppose we are interested in the effect of a subset ${\bf X}_{S}$ (where the subscript ``S'' refers to a subset of features in ${\bf X}$) on the output of $g(x)$. The partial dependence
of $g(x)$ on ${\bf X}_{S}$ is defined as:
\begin{equation}
    g_{S}(x_{S})=\mathbb{{E}}_{\mathbf{X}_{C}}[g(x_{S},\mathbf{X}_{C})]=\int g(x_{S},x_{C})dP(x_{C}),
\end{equation}
where $\mathbf{X}_{C}=\mathbf{X}\setminus\mathbf{X}_{S}$. PDP shows the marginal effect of a feature or a set of features on the predictions. Visualizing PDP is feasible for only one or two features at a time. The recent work of \cite{zhao2019causal} has shown that when a causal structure of the domain is available, PDP  can be used, under some conditions, to measure the causal effect of some feature(s) on the prediction. However, such a causal interpretation of PDP requires a structural causal model to determine whether a particular subset of features satisfy the necessary conditions for causal interpretation. It would be interesting to explore the relationship between PDP and its extensions, e.g., individual conditional expectation \cite{goldstein2015peeking}, and the causal attribution of predictions to the features introduced in this paper.  

It is worth noting that there is a body of recent work on algorithmic fairness \cite{khademi2019fairness,khademi2019algorithmic,kusner2017counterfactual,zhang2018equality,huan2020fairness,nabi2018fair}  which has shown that the problem of determining whether or not a predictive model is discriminatory with respect to a protected attribute, e.g., gender, race, can be reduced to the problem of determining whether the protected attribute has a causal effect on the output of the predictive model.  A predictive model is deemed non-discriminatory with respect to a protected attribute if the causal effect of the protected attribute on the output of the predictive model is negligible (ideally 0). Clearly, methods such as those introduced in this paper for black box predictive model interpretation via causal attribution can be applied to answer algorithmic fairness questions, and help ensure that such models do not become instruments of unfair discrimination on the basis of gender, race, etc. We further note that most of the work on causal criteria for algorithmic fairness have focused on protected attributes that are  binary or categorical. The causal attribution method introduced in this paper can cope with protected attributes that are ordinal or continuous e.g., age. 

\subsection{Limitations and Caveats}
The proposed approach to causal attribution relies on state-of-the-art methods for estimation of causal effects from observational data, and as such, the obtained causal attributions depend on the estimated causal effects. These estimates, in turn, depend on the accuracy of the specific method used to estimate the causal effect of each input of the predictive model on the model's output. Furthermore, sometimes the different models disagree on the magnitude as well as the sign of the estimated causal effect. Under the circumstances, in practice, it makes sense to trust the causal attributions where the corresponding causal effects estimated by the different methods are largely in agreement with each other, and question the causal attributions where the corresponding causal effect estimates disagree with each other. 

The proposed approach to black box predictive model interpretation via causal attribution assumes that the causal effect of each of the model inputs on the model output can in fact be estimated using only the observations of the model's input-output behavior on a sufficiently large number of data samples. Estimation of causal effects from observational data within the potential outcomes framework relies on the  strong ignorability  assumption  which can be roughly paraphrased as saying that any causal relationships between the potential outcomes and the treatment are fully explained by the observables, i.e., that there are no unmeasured confounders \cite{hernan2020causal}. Strong ignorability is the counterpart of the back-door criterion  \cite{pearl2010foundations}, a graphical criterion that if satisfied by a set of observed variables ${\bf Z}$ of a structural causal model, is a sufficient condition for identifiability of the causal effect of a treatment $T$ on an outcome $Y$ from observational data. It requires that no element of ${\bf Z}$ is a graphical descendent of $T$ and ${\bf Z}$ blocks all {\em back door} paths from $T$ to $Y$. Such a set ${\bf Z}$ of observed variables correspond precisely to the set of confounders that need to be controlled for in estimating the causal effect of $T$ on $Y$ from observational data. However, the backdoor criterion, or equivalently, strong ignorability, cannot be verified or refuted from {\em only} observational data \cite{pearl2009causality}, and requires a structural causal model. In general, such a structural causal model has to be either assumed, perhaps based on prior knowledge, or derived, e.g., by translating a deep neural network or other predictive model trained on an observational data set \cite{chattopadhyay2019neural}. In either case, the validity of such a structural causal model cannot be verified from observational data alone. 

It is possible to cope with violations of the strong ignorability assumption in some settings. For example, one can cope with {\em unmeasured confounders}, whenever possible, by identifying the confounders (using background knowledge) and adjusting for them \cite{hernan2020causal,pearl2009causality}. A second approach is to use instrumental variables  \cite{angrist1996identification}. A third approach is to estimate bounds on the causal effects  instead of the causal effects themselves \cite{manski2009identification}. Development of methods that cope with violations of the strong ignorability assumption in the presence of hidden confounders remains an active area of research \cite{finkelstein2020deriving}.


\subsection{Future Work}
Causal attribution is a small, albeit important and necessary step towards explaining complex predictive models trained using machine learning. However, in addition to being causally grounded, explanations have to be selective, formulated at the appropriate level of abstraction, context-specific, concise, and framed relative to the knowledge  of the explainee and the purpose of the explanation. Almost all of the work on explaining predictive models trained using machine learning fall short of these objectives. Hence, there is much room for (i) developing more advanced approaches to producing explanations from predictive models, data, background knowledge and assumptions, context, etc.; (ii) relaxing some of the limiting assumptions of the potential outcomes framework to broaden the scope of applicability of the causal attribution methods introduced in this paper; (iii) developing better criteria for evaluating and comparing alternative explanations; and (iv) approaches to integrating explanatory machinery into robust, interactive, transparent, accountable, and trustworthy human-AI systems.

\subsubsection*{Acknowledgements}
This work was funded in part by the  NIH NCATS through the grant UL1 TR002014 and by the NSF through the grants 1636795 and 2041759, the Edward Frymoyer Endowed Professorship  at Pennsylvania State and the Sudha Murty Distinguished Visiting Chair in Neurocomputing and Data Science funded by the Pratiksha Trust at the Indian Institute of Science (both held by Vasant Honavar). The content is solely the responsibility of the authors and does not necessarily represent the official views of the sponsors.

\bibliographystyle{abbrv}
\bibliography{bibliography.bib}

\end{document}